\documentclass[10pt,journal,compsoc]{IEEEtran}

\usepackage{amsmath,amsfonts}
\usepackage{algorithmic}
\usepackage{algorithm}
\usepackage{array}
\usepackage[caption=false,font=normalsize,labelfont=sf,textfont=sf]{subfig}
\usepackage{textcomp}
\usepackage{stfloats}
\usepackage{url}
\usepackage{verbatim}
\usepackage{graphicx}
\usepackage{cite}
\usepackage{pifont}
\usepackage{xcolor}
\usepackage{booktabs}
\usepackage{multirow}
\usepackage[switch]{lineno}
\usepackage{enumitem}

\definecolor{ACMPurple}{cmyk}{0.55,1,0,0.15}
\definecolor{ACMDarkBlue}{cmyk}{1,0.58,0,0.21}

\usepackage[colorlinks,linkcolor=purple,citecolor=violet,anchorcolor=blue,urlcolor=blue]{hyperref}

\newcommand{\cmark}{\ding{51}}%
\newcommand{\xmark}{\ding{55}}%

\begin{document}

\title{Federated Continual Learning via Knowledge Fusion: A Survey}

\author{Xin Yang,~\IEEEmembership{Member, IEEE}, Hao Yu, Xin Gao, Hao Wang, \\
Junbo Zhang,~\IEEEmembership{Senior Member, IEEE}, Tianrui Li,~\IEEEmembership{Senior Member, IEEE}
\thanks{Xin Yang, Hao Yu and Xin Gao are with the School of Comput-
ing and Artificial Intelligence, Southwestern University of Finance and Economics, Chengdu, 611130, China.
E-mail: {yangxin@swufe.edu.cn, yuhao2033@163.com, xingaocs@foxmail.com}.}
\thanks{Hao Wang is with Nanyang Technological University, Singapore.
E-mail: {cshaowang@gmail.com} (primary) or {h.wang@ntu.edu.sg}.
}
\thanks{Junbo Zhang is with JD iCity, JD Technology, Beijing, China, JD Intelligent Cities Research \& Institute of Artificial Intelligence, Southwest
Jiaotong University, Chengdu 611756, China.
E-mail: {msjunbozhang@outlook.com}.}
\thanks{Tianrui Li is with School of Computing and Artificial Intelligence, Southwest Jiaotong University, Chengdu, 611756, China. 
E-mail:{trli@swjtu.edu.cn}.}
}



\IEEEtitleabstractindextext{%
\begin{abstract}
Data privacy and silos are nontrivial and greatly challenging in many real-world applications. Federated learning is a decentralized approach to training models across multiple local clients without the exchange of raw data from client devices to global servers. 
However, existing works focus on a static data environment and ignore continual learning from streaming data with incremental tasks. 
Federated Continual Learning (FCL) is an emerging paradigm to address model learning in both federated and continual learning environments. The key objective of FCL is to fuse heterogeneous knowledge from different clients and retain knowledge of previous tasks while learning on new ones.
In this work, we delineate federated learning and continual learning first and then discuss their integration, i.e., FCL, and particular FCL via knowledge fusion. In summary, our motivations are four-fold: we (1) raise a fundamental problem called ``spatial-temporal catastrophic forgetting'' and evaluate its impact on the performance using a well-known method called federated averaging (FedAvg), (2) integrate most of the existing FCL methods into two generic frameworks, namely synchronous FCL and asynchronous FCL, (3) categorize a large number of methods according to the mechanism involved in knowledge fusion, and finally (4) showcase an outlook on the future work of FCL.

\end{abstract}

}

\maketitle
\section{Introduction}
Deep learning has been instrumental in the advances of both data analysis and artificial intelligence (AI) \cite{lecun2015deep}. The algorithms have also been successfully used in almost all areas of applications in industry, science, and engineering \cite{dong2016comparison}. However, traditional deep learning algorithms often require large amounts of training data and centralized training, which are usually impractical or even impossible in practical situations. For example, in the case of mobile devices, it is difficult to collect and transfer large amounts of data to a central server with limited bandwidth \cite{mothukuri2021survey}. Centralized training is also infeasible due to data privacy and data silos because of the information security, related laws and regulations, and intensive competition among technology giants \cite{zhang2021survey}. In recent years, federated learning (FL) has emerged as a positive response to the increasing need for privacy and other concerns in machine learning applications \cite{aledhari2020federated}. This newly emerging paradigm deploys multiple devices to train a global model collaboratively by uploading their models to a server while keeping the private data locally, as opposed to the traditional centralized data storage and model training approach. FL mainly involves privacy preservation, knowledge transfer, and collaboration among data owners, which are essential for many real-world applications of AI. After several years of development, FL has produced promising results, particularly in industries that handle sensitive data such as finance, healthcare, electronic communication and government affairs \cite{rieke2020future,kang2020reliable,brisimi2018federated}.

Despite its numerous advantages, FL confronts significant challenges. The first challenge is the lack of dynamism. The setting assumptions of most existing FL methods are somewhat strong, assuming all data and classes should be known prior and will be unchanged forever. It contradicts the dynamic nature of the real world, where data are collected in streams and the data could form sequential tasks. 
However, traditional FL would suffer from terribly temporal catastrophic forgetting (\textit{TemporalCF}) when local models learning on new tasks. \textit{TemporalCF} is a major challenge in Continual Learning (CL) \cite{diethe2019continual}, which refers to changes in crucial parameters of a single model when learning on consecutive tasks, resulting in poor performance on previous tasks \cite{mccloskey1989catastrophic}. Apparently, FL would certainly encounter \textit{TemporalCF} if each client learns on a sequence of tasks from a private local data stream.


Another challenge of FL is data heterogeneity \cite{zhang2021survey}, also known as non-independent and identically distributed (Non-IID) data. It refers to the differences in data distributions, feature spaces and labels across clients participating in the federated training process, aroused by variations in the data sources, data collection methods, data quality, etc. Data heterogeneity results in diverse knowledge extracted by local models. Knowledge of a model is often represented by parameters. The more different the data, the greater the parameter divergence. Simply aggregating these local models, like averaging parameters, can lead to critical parameters for certain local tasks being overwritten. Such that the aggregated global model may perform worse than the local models on the local test set \cite{ghosh2019robust,luo2022tackling,zhu2021data}. We call this phenomenon spatial catastrophic forgetting (\textit{SpatialCF}). It is worth noting that \textit{SpatialCF} is a new concept we introduce in this paper, which is analogical to the \textit{TemporalCF}.

Federated Continual Learning (FCL) \cite{wang2019adaptive,kairouz2021distributed} breaks the static limitations of traditional federated learning on each client learning a task sequence. Beyond FL and CL, FCL has a fundamental problem called \textit{spatial-temporal catastrophic forgetting}. On the local side, clients need to overcome \textit{TemporalCF} induced by learning new tasks. On the server side, the server must address \textit{SpatialCF} caused by aggregating different local models. Specifically, after completing the training of a task, the server aggregates local models into a global model. Then the clients use the global model as a foundational model for continual learning of the next task. Therefore, \textit{SpatialCF} may exacerbate \textit{TemporalCF}.


In this paper, we enumerate several FCL algorithms profoundly to explore how they overcome spatial/temporal catastrophic forgetting or both. We evaluate them with extensive experiments and show that the training process of FCL is essentially a sort of knowledge fusion: fuse knowledge of previous tasks with current task's knowledge, as well as fuse heterogeneous knowledge from different clients. Specifically speaking, since uploading the entire local parameters or gradients to the server may cause privacy leakages \cite{bhagoji2018model,zhu2019deep} and slow down the aggregation due to the large communication cost \cite{luping2019cmfl,lin2017deep}, FCL methods typically extract knowledge during local training, upload the extraction of local knowledge and explore more efficient model aggregation strategies.

The main contributions of this paper are summarized as follows:
\begin{itemize}[leftmargin=*] 
    \item We discuss FCL thoroughly in this work. We raise the problem of spatial-temporal CF. We further define two evaluation metrics and conduct extensive experiments to evaluate the effect of spatial-temporal CF on FCL.
    \item We propose two unified frameworks for FCL (i.e., synchronous and asynchronous FCL). The two frameworks can integrate most existing methods and lay two pipelines for future research.
    \item We showcase a comprehensive survey of existing FCL methods by categorizing them with seven forms of knowledge fusion, namely rehearsal, clustering, regularization, parameter isolation, dynamic architecture, prototype and knowledge distillation.
\end{itemize}

The rest of this paper is organized as follows. In Sec.~\ref{Sec2}, we provide an overview of FL and CL, and discuss the motivation for their integration, i.e., FCL. In Sec.~\ref{Sec3}, we introduce the problem of spatial-temporal catastrophic forgetting and demonstrate the weaknesses of traditional FL for this problem through experiments. In Sec.~\ref{Sec4}, we present two generic FCL frameworks: synchronous FCL and asynchronous FCL. In Sec.~\ref{Sec5}, we delve into various FCL methods and categorize them according to the forms of knowledge fusion, namely \textit{rehearsal}, \textit{clustering}, \textit{all gradients/parameters}, \textit{parameter isolation}, \textit{prototype}, \textit{dynamic architecture} and \textit{knowledge distillation}. In Sec.~\ref{Sec6}, we briefly summarize our survey and highlight some potential research directions for future work in FCL.

\section{Federated and Continual Learning: An Overview}
\label{Sec2}
\subsection{Federated Learning}
In recent years, the rapid development of network technology and AI has brought about a growing concern regarding the potential risks of privacy disclosure. As citizens and companies become increasingly aware of personal privacy and data ownership, governments have taken measures to address privacy risks and security threats \cite{lyu2020threats}. Notably, the European Union issued the General Data Protection Regulation \cite{regulation2018general} in 2018, which represents the first bill on data privacy protection. In May 2019,  San Francisco banned the use of face recognition by government agencies to eliminate the hidden dangers caused by technology \cite{zeng2019responsible}. In addition to laws and regulations, information asymmetry, an important factor in improving competitiveness, is also against the needs of traditional deep learning for centralized data storage and processing. On the one hand, many data are of poor quality and lack labels, while on the other hand, data are scattered across various data subjects and enterprises, creating data silos that cannot be connected \cite{yang2020federated}. In brief, there is an insoluble dilemma of protecting data privacy and breaking data silos in the traditional centralized machine learning paradigm.

FL has been proposed as a potential solution to this dilemma \cite{9599369}. FL is a distributed learning approach where multiple devices or nodes collaboratively train a shared model without sharing their raw data \cite{yang2020federated}. It originated from FedAvg \cite{Communication-efficientlearningofdeepnetworksfromdecentralizeddata}, which allows each client to use local data to train their own local model, then local models will be aggregated as a global model by weighted average based on the proportion of data, the global model will then be distributed to clients who take part in aggregation and help to train a new initial model in the next round of communication. The core idea of FL is to break down data silos while preventing privacy breaches by fusing extracted knowledge, instead of raw data, of edge devices \cite{yang2019federated}.

Since then, various FL algorithms have been proposed. However, recent studies have evidently shown that uploading full gradients to the server fails to protect the raw data. \cite{zhu2019deep} proposes Deep Leakage from Gradients (DLG), an algorithm that can obtain original local data from publicly shared gradients.\cite{zhao2020idlg} shows that the ground-truth labels can be extracted with 100\% accuracy under DLG attack. \cite{geiping2020inverting} demonstrates the feasibility of reconstructing images at high resolution from the knowledge of uploaded parameter gradients. Other forms of attacks also pose threats to the security and privacy of FL. Attacks conducted during the inference phase are called evasion or exploratory attacks \cite{lyu2022privacy}. \cite{hitaj2017deep} firstly devised an active inference attack called Generative Adversarial Networks (GAN) attack against deep FL models. It allows the adversarial participant to train a GAN to generate prototypical samples of the targeted private training data.

To further enhance privacy protection, researchers have explored alternative information forms to exchange instead of original gradients. We summarize alternative approaches in FCL settings in the later section.

According to the different distribution characteristics of data, existing FL methods can be categorized into horizontal federated learning (HFL), vertical federated learning (VFL), and federated transfer learning (FTL) \cite{yang2019federated,yang2020federated}. HFL refers to the case where clients share a similar feature space but differ in sample space, which occurs in cross-silo and cross-device scenarios. VFL refers to the case where clients share a similar sample space but differ in feature space, which typically occurs in the cross-silo scenario. FTL refers to the situation that neither overlaps of feature space nor sample space are large, requiring some transfer learning approaches to assist FL training.

It is worth mentioning that although FL has achieved remarkable success in addressing the data silo issues and privacy concerns, and it is recognized as a promising direction, data heterogeneity has become a major bottleneck of this paradigm.  Data heterogeneity is an inherent challenge in FL, as the data distribution across different clients is typically Non-IID due to differences in data sources, data collection methods, and data quality.  Local models trained with heterogeneous data are also heterogeneous, which leads to poor performance when aggregating these models to produce a global model. After distributing it to clients, its performance may not be as good as some local models \cite{yu2020salvaging}. We coin the term ``\textit{spatialCF}'' to describe this phenomenon, since its intrinsic similarity with the scenario of CL--only a portion of data is available. Specifically, catastrophic forgetting in CL is due to the temporal unavailability of data, while heterogeneity in FL is caused by the spatial unavailability of data.

\subsection{Continual Learning}

In reality, the entire dataset is not available all at once but rather arrives continually in the form of task streams. The sequential training on new tasks often results in overwriting parameters of the model, leading to a significant decline in performance on previous tasks \cite{parisi2019continual, kemker2018measuring,kim2022theoretical}. The goal of CL is to alleviate forgetting by integrating the knowledge learned from previous tasks and the current task \cite{hadsell2020embracing,ke2020continual}. 

Several techniques have been proposed to address the challenges of CL, including regularization, parameter separation, dynamic architecture, replay, and knowledge distillation \cite{mai2022online}. Regularization techniques, such as Elastic Weight Consolidation (EWC) \cite{kirkpatrick2017overcoming}, aim to constrain the changing of important parameters to prevent catastrophic forgetting. Parameter separation techniques, such as PackNet \cite{mallya2018packnet} and PathNet \cite{fernando2017pathnet}, only activate relevant parameters for a certain task and keep the network the same in scale. Dynamic architecture methods, such as Progressive Neural Networks (PNNs) \cite{rusu2016progressive} and Expert Gate \cite{aljundi2017expert},  add parameters
for new tasks while leaving the old parameters unchanged. Some studies \cite{de2021continual} consider parameter separation and dynamic architecture techniques to be in the same category, namely parameter isolation, as they both reduce interference between new and old tasks by isolating parameters. The only difference is whether the structure of the network is dynamic or fixed. Replay-based methods, such as Experience Replay (ER) \cite{rolnick2019experience} and Generative Replay (GR) \cite{shin2017continual}, store and replay previously seen data to mitigate forgetting. Knowledge distillation methods, such as Knowledge Transfer via Distillation (KTD) \cite{heo2019knowledge} and Learning without Forgetting (LwF) \cite{li2017learning}, transfer knowledge from previously learned models to new models periodically to mitigate forgetting.

We believe the primary challenge faced by FL is comparable to CL: integrating knowledge learned from multiple clients' data. The purpose of CL is to integrate knowledge from different time into a single model, conducted sequentially. In contrast, FL aims to fuse knowledge from multiple clients into one global model, carried out in parallel. Both CL and FL share the goal of fusing diverse knowledge. To some extent, FL can be seen as a sort of traditional CL but across different machines. Evidently, CL methods would be helpful in resolving FL problems.

Based on the analysis above, many researchers have tried to adapt CL methods to FL settings and achieved quite good results. Shoham $et\, al.$ \cite{shoham2019overcoming} presents their adaptation of the EWC \cite{kirkpatrick2017overcoming}, a classic regularization-based approach in CL, to the FL scenario and refers to it as FedCurv. By adding a penalty term to the loss function, all local models are compelled to converge toward a shared optimal. Xin $et\, al.$ propose a new framework called FedCL \cite{yao2020continual}. FedCL follows the parameter-regularization continual training paradigm on clients, in other words, penalizing the important parameters of the global model for changing. Nevertheless, it is quite different from FedCurv because it estimates the importance weights on the proxy dataset on the server and then distributes them to the clients instead of estimating the importance weight of model parameters on clients and exchanging them as in \cite{shoham2019overcoming}, which will bring at least twice the extra communication costs.

\textit{SpatialCF} in FL and the close relationship between FL and CL are also observed and demonstrated by other researchers. Casado $et\, al.$ \cite{casado2021concept} creatively modifies the traditional FedAvg algorithm by adopting the concept detection and concept drift adaptation methods that deal with concept drift in CL and prove that their extended method outperforms the original one. Criado $et\, al.$ \cite{criado2022non} dig deeply the connection between concept drift in CL and Non-IID in FL, and provide an enlightening survey on handling Non-IID problem. Usmanova $et\, al.$ \cite{usmanova2022federated}  shows that the problem of catastrophic forgetting is critical in a pervasive computing application using FL. It is evident that the close relationship between CL and FL has been confirmed.

After years of development, research on overcoming catastrophic forgetting in FL has been increasingly in-depth \cite{ijcai2022p0303.,bui2018partitioned,yoon2021federated}. Since data do not arrive at each client at once, there is also \textit{TemporalCF}. Besides, the task sequences of clients are heterogeneous and the time of uploading extracted knowledge to the server is diverse, making the problem even more complicated \cite{li2020federated}. Therefore, both local and global models should be capable of fusing knowledge continually. FCL is proposed to address this issue. Adding CL to traditional FL can reduce the cost of data storage and model retraining when adding classes and reduce the risk of data leakage due to model gradient updates.

\section{Spatial-Temporal Catastrophic Forgetting}
\label{Sec3}
\subsection{Problem Definition}
FCL suffers from catastrophic forgetting both in terms of time and space. On the one hand, the local model should have the ability to preserve previous knowledge while learning from the newest data collected progressively. On the other hand, the global model, which is obtained by aggregating local models, often performs worse than local models on the local dataset due to Non-IID data. 

Our research now finds that both types of catastrophic forgetting are actually caused by the unavailability of the entire dataset. In CL, the arrival of new data renders old data inaccessible, resulting in \textit{TemporalCF}. In FL, the unavailability arises due to the distributed storage of data \cite{wei2022knowledge}, leading to \textit{SpatialCF}.

\begin{table}[htbp] 
\centering
\caption{Mathematical Notations.}
\label{notation}
\vspace {-2.5mm}
\begin{tabular}{cl} 
\hline
\ Notation & \quad \quad \quad \quad Description\\
\hline
 $S$ & Central server of FL system.\\
 $C$ & Total number of clients.\\
 $\mathcal{T}^c$ & Entire task sequence of client $c$.\\
 $\mathcal{T}^c_t$ & $t$-th task on client $c$.\\
 $T$ & Total number of tasks of all clients.\\
 $N_c$ & Number of tasks in the task sequence of client $c$.\\
 $r$ &Communication round.\\
 $\theta_g^r$ & Global model at round $r$.\\
 $\theta_c^r$ & Local model of client $c$ at round $r$.\\
 $O_c$ & Order of tasks on client $c$.\\
 $A_i^k$ & Class set of Client $i$ on its \textit{k-th} task.\\
 $KR_t$ & Temporal knowledge retention.\\
 $KR_s$ & Spatial knowledge retention.\\
\hline 
\end{tabular}
\end{table}

The mathematical notations involved are shown in Tab.~\ref{notation}. For client $c \in C$, where $C$ denotes the total number of clients in FCL system, the local model $\theta_c$ is trained on its private task sequence  $\mathcal{T}^c=\{\mathcal{T}^c_1,\mathcal{T}^c_2,\dots,\mathcal{T}^c_{N_c}\}$ . The objective of FCL is to train a generalized global model, which effectively avoids spatial-temporal forgetting for previous tasks of all clients. Formally, we measure the forgetting of the global model of $s$ communication round $\theta_g^s$ on the task $t$ presented at round $r$ by the cross-entropy loss 
\begin{equation}
    \mathcal{L}_{CE}(\theta_g^r,\theta_g^s)=-\sum^{n_t}_{i=1}\theta_g^r(x_i)log\theta_g^s(x_i),
\end{equation}
where $x_i\in[1,n_t]$ denotes the samples of task $t$. And the spatial-temporal forgetting of all seen tasks is defined as follows. The spatial-temporal forgetting at round $s$ is the average loss of  all seen tasks:
\begin{equation}
    F(s,T)=-\frac{1}{T}\sum^T_{t=1}\sum^{n_t}_{i=1}\theta_g^r(x_i)log\theta_g^s(x_i),
\end{equation}
where $T$ denotes the total number of all seen tasks at round $s$.
We then verify that spatial-temporal catastrophic forgetting does exist and has a great impact on the performance of both local and global models. Moreover, these experiments also aim to identify the most significant factor that leads to the performance degradation of the models.

\subsection{Performance Evaluation}
\subsubsection{Settings}
\label{sec:3.2.1}

We conduct experiments using the classical FL algorithm FedAvg in four scenarios where the data distribution varies among clients. The aim is to explore the key factors leading to spatial-temporal catastrophic forgetting. The data distribution scenarios between clients are shown as follows: 
\begin{enumerate}[label=(\Alph*)]
\item The clients share the same classes, and their data distributions are consistent. In this scenario, each client is allowed to have the same set of data classes, and the number of samples for each class is identical across all clients.
\item The clients share the same classes, while their data distributions are different. This aligns with the previous scenario, except that the number of samples for each class varies across different clients.
\item The clients have different sets of classes, while there are overlapping classes among them. It allows each client to have different classes, but there are still some classes that are common to all clients, referred to as overlapping classes.
\item The clients have different sets of classes. Each client has a very unique set of classes and there is no overlapping class between clients. 
\end{enumerate}

In each scenario, we further divide the task sequence handled by each client into three different settings:
\begin{enumerate}[label=(\alph*)]
    \item IID: Each task includes all the classes, and the number of samples for each class in each task is equal.
    \item Non-IID: Each task includes all the classes, but the number of samples for each class in each task is not equal.
    \item Class-Incremental: Each task evenly distributes all classes, with each class appearing in only one task. For example, if there are a total of 30 classes and the client has 5 tasks, each task will have data for only 6 classes.
\end{enumerate}

Tab.~\ref{tab1} illustrates our experimental settings. For our model, we set the first convolution layer in the Res-Net 18 \cite{he2016deep} to 3$\times$3 in experiments.

\begin{table}[htbp]
\caption{Experimental setting in Section 3.2.}
\label{tab1}
\vspace {-7mm}
\begin{center}
\resizebox{0.5\textwidth}{!}{
\begin{tabular}{ccccc}
\hline
&\multicolumn{3}{c}{Across Clients}&Across Tasks\\
\hline
&Classes&Distribution&Overlapping classes&Distribution\\
\hline
$a$&Same&IID&/&IID\\
$b$&Same&IID&/&Non-IID\\
$c$&Same&IID&/&Class-Incremental\\

$d$&Same&Non-IID&/&IID\\
$e$&Same&Non-IID&/&Non-IID\\
$f$&Same&Non-IID&/&Class-Incremental\\

$g$&Different&/&\cmark&IID\\
$h$&Different&/&\cmark&Non-IID\\
$i$&Different&/&\cmark&Class-Incremental\\

$j$&Different&/&\xmark&IID\\
$k$&Different&/&\xmark&Non-IID\\
$l$&Different&/&\xmark&Class-Incremental\\
\hline
\vspace{-5mm}
\end{tabular}}
\end{center}
\end{table}

\subsubsection{Datasets}
We conduct twelve experiments on CIFAR-100 \cite{krizhevsky2009learning} in an FL system that has three clients and one server. Every client would be trained on three tasks sequentially. 
Each task's data is split into a training set and a test set in a ratio of 7:3. Every client maintains a large test set that consists of the test dataset from the seen tasks so far. Therefore, the evaluation of the current model and the aggregated global model will run on this large test set.

For each scenario of data distribution among clients, the designed data distribution is different, as shown below:
In scenario \textbf{(A)} and \textbf{(B)}, the initial step involves randomly selecting 30 classes from a pool of 100. In Scenario \textbf{(A)}, the data of each class is evenly distributed among three clients, ensuring an equal quantity of data for each client. Conversely, in Scenario \textbf{(B)}, the data of each class is randomly assigned to different clients using a Dirichlet distribution. While all clients have data for every class, the sample quantities for each class are distributed randomly. In scenario \textbf{(C)}, each client initially randomly samples 15 classes exclusive to itself. Subsequently, from the remaining classes, 15 overlapping classes are selected. The data for these overlapping classes is evenly distributed among all clients. As for scenario \textbf{(D)}, there are no overlapping classes among clients. Therefore, each client possesses 30 classes exclusively owned by itself.

According to Sec.~\ref{sec:3.2.1}, there are three further setting for each scenario and the detailed descriptions for each setting are as below: \textbf{(IID)} means that the sample quantity for each class is the same across all tasks, and there are no repeated samples. And \textbf{(Non-IID)} implies that the sample quantity for each class varies across different tasks, determined by a Dirichlet distribution. \textbf{(Class-Incremental)} signifies that each task contains data for only 10 classes, and there is no repetition of classes across tasks.

The objective of our experiments is to identify the major factors causing the performance degradation of federated learning models.

\begin{figure*}[htbp]
	\centering
	\includegraphics[width=0.7\textwidth]{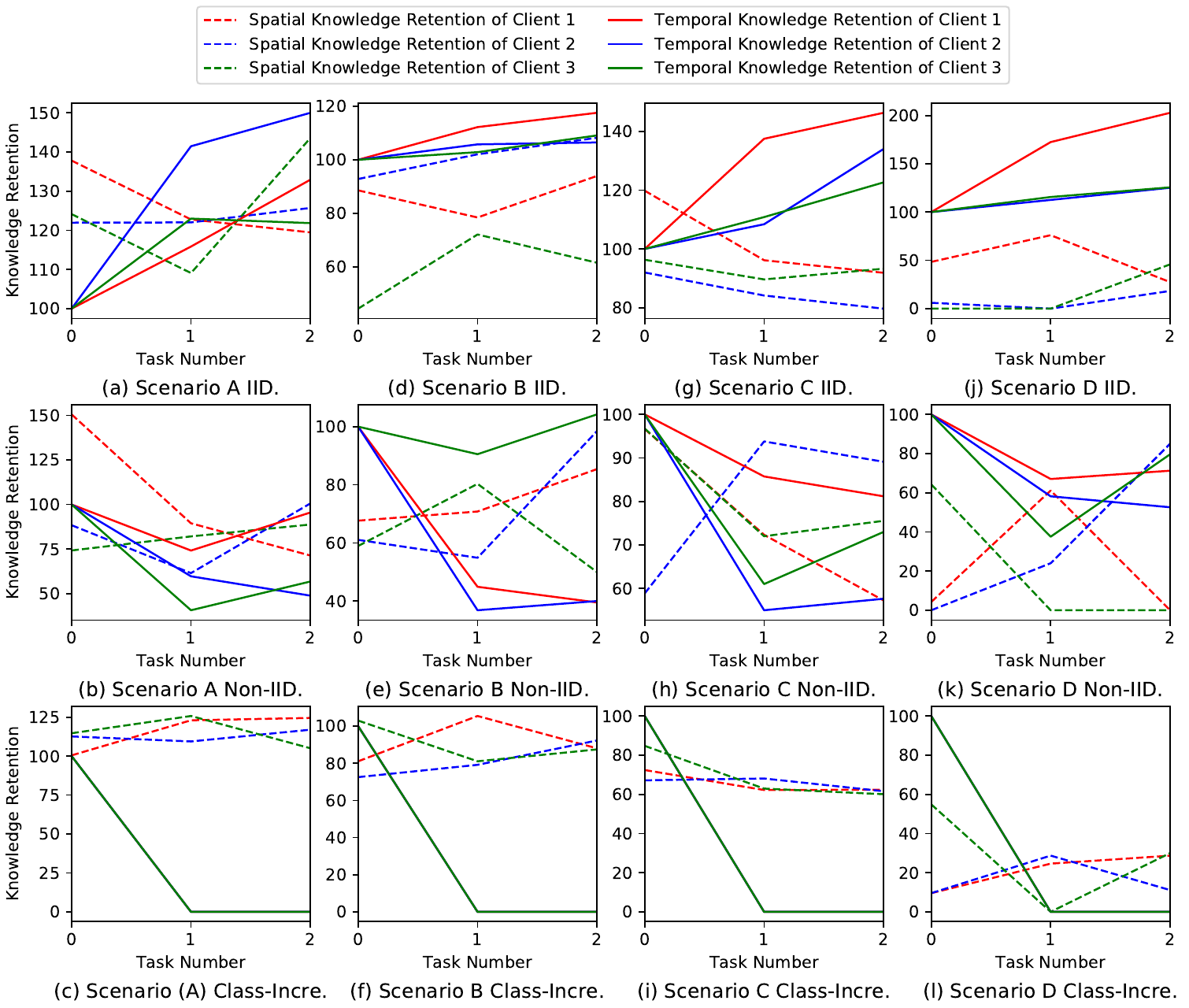}
  \caption{Experiment Results. Solid lines represent knowledge retention of clients in time, and dashed lines represent knowledge retention of clients in space. When the data distribution among clients is identical, and the distribution among local tasks is also the same, spatial-temporal catastrophic forgetting does not occur, and the model's performance improves. However, severe performance degradation occurs after aggregation when clients have training data with different classes. Additionally, when the local task sequence follows a class-incremental pattern, the model forgets all knowledge from previous tasks, resulting in temporal knowledge retention to zero.}
  \label{fig:results}
\end{figure*}
\subsubsection{Metrics}
Since spatial-temporal catastrophic forgetting is a novel challenge that we first introduced, lacking measurements, we have designed three different metrics to assess temporal knowledge retention and spatial knowledge retention.

\textbf{Temporal knowledge retention.} We use $Knowledge$ $Retention$ as measurement of forgetting. Temporal knowledge retention is designed to measure the extent to which local models retain knowledge of old tasks as they learn on the task sequence. ${Acc}_i^{(0,0)}$ represents the accuracy of client $i$'s local model trained on the first task testing on the test set of the first task. And ${Acc}_i^{(r,0)}$ represents the accuracy of client $i$'s local model trained on the $r$-th task testing on the test set of the first task. The ratio of these two values provides insight into how much knowledge the local model retains from the first task when it has completed training on the $r$-th task. Therefore, the spatial catastrophic forgetting can be expressed in Equ.~\ref{krt}.
\begin{equation}
\label{krt}
    KR_t = \frac{{Acc}_i^{(r,0)}}{{Acc}_i^{(0,0)}}
\end{equation}

\textbf{Spatial knowledge retention.} Similarly, we can deduce the expression form of spatial catastrophic forgetting in Equ.~\ref{krs}. This metric is designed to measure how much local-specific knowledge is retained by the aggregated global model. A smaller value indicates that more local knowledge was overwritten during aggregation.
\begin{equation}
\label{krs}
    KR_s = \frac{{Acc}_g^{(r,r_i)}}{{Acc}_i^{(r,r)}}
\end{equation}
where ${Acc}_g^{(r,r_i)}$ is the accuracy of the global models on the $r$-th testset of client $i$. And the global model is obtained by aggregating the local models trained on the $r$-th task from all the clients.

\subsubsection{Experimental Results and Analysis}
At the beginning of the experiment, each client has a task sequence of three different tasks. After completing the training for a task, they will upload it to the server and participate in aggregation. Subsequently, they employ the received global model for continual learning on the next task.

In Fig.~\ref{fig:results}, each row represents a type of data distribution across tasks, such as IID. Meanwhile, each column represents a scenario among clients, e.g., Scenario \textbf{A}. From the last row of Fig.~\ref{fig:results}, we can tell that when the task sequence is class-incremental, traditional FL methods have no ability to alleviate \textit{TemporalCF}. On the contrary, if the distributions among tasks are entirely identical, the local model will consistently learn from similar data, continually enriching its knowledge. In this scenario, the model's performance on the initial test set may significantly improve, as it can accumulate and leverage similar patterns and features during the training process.

As for \textit{SpatialCF}, we observed that Scenario \textbf{A} exhibits the highest spatial knowledge retention, especially when the distribution among tasks is class-incremental. Conversely, Scenario \textbf{D} has the lowest spatial knowledge retention, particularly in the setting where tasks follow an IID distribution. When clients share the same classes and each client has an equal number of samples, local models become more similar, leading to enhanced performance after weighted averaging. The Class-Incremental setting reduces the complexity of each task, further minimizing the differences among local models. However, when each client has different classes, the heterogeneity among local models intensifies, causing crucial parameters to be overwritten during aggregation, resulting in a sharp decline in performance on local tasks.

Based on the analysis, we can draw a very important conclusion: \textbf{The divergence in distributions among clients leads to spatial forgetting, whereas differences in distributions among tasks result in temporal forgetting. }The greater the variance in data among clients, such as differences in classes, the more crucial it becomes to design an effective knowledge fusion algorithm for amalgamating heterogeneous models, rather than a straightforward aggregation approach. A good FL algorithm must be able to effectively integrate knowledge extracted from various clients, especially retaining knowledge of seen classes to prevent performance degradation caused by catastrophic forgetting.


\section{Federated Continual Learning}\label{Sec4}
After the above analysis, the objective of FCL is to address spatial-temporal catastrophic forgetting due to the application of FL in the real world. The first is to ensure that local models do not forget the knowledge of previous tasks while learning new ones, which refers to overcoming temporal catastrophic forgetting. The second is to enable the global model to fuse all the heterogeneous knowledge across different clients without performance degradation on local test set , addressing the problem of spatial catastrophic forgetting.

In traditional FL setting \cite{yang2019federated,9599369},  
there are $c$ clients ${C_{1}, \dots, C_c}$ and one central server $S$. And client $C_i, 1\leq i\leq c$ only has access to its own data $D_i$ due to privacy concerns. Basically, there are three steps in one communication round: (1) Server $S$ distributes the initial model or the global model from the last round to clients. (2) Client $C_i$ would use its private data $D_i$ to train its local model $M_i$ based on the model from the server. (3) Server collects local model ${M_1, \dots,M_c}$ then aggregates them to update the global model. The performance of the final global model should be very close to the performance of a centralized trained model.

We then extend conventional FL to federated continual learning. Given $c$ local clients, each of them $\{C_1,\dots,C_c\}$ trains their local model on a private task sequence $T_i=\{t_i^1,\dots,t_c^{i_i}\}$, where $n_i$ represents the total number of tasks in this task sequence. In other words, in FCL the system needs to train $c$ continual learning model with their task sequence. And the server will collect these local CL models and fuse them to form a global model, then distribute it to the clients. 

Obviously, the biggest challenge for FCL is to overcome the temporal forgetting caused by learning on a data stream, and the spatial forgetting happens when aggregating different local models on the server.

\subsection{Synchronous FCL}
\begin{figure*}[h]
  \centering
  \includegraphics[width=0.8\textwidth]{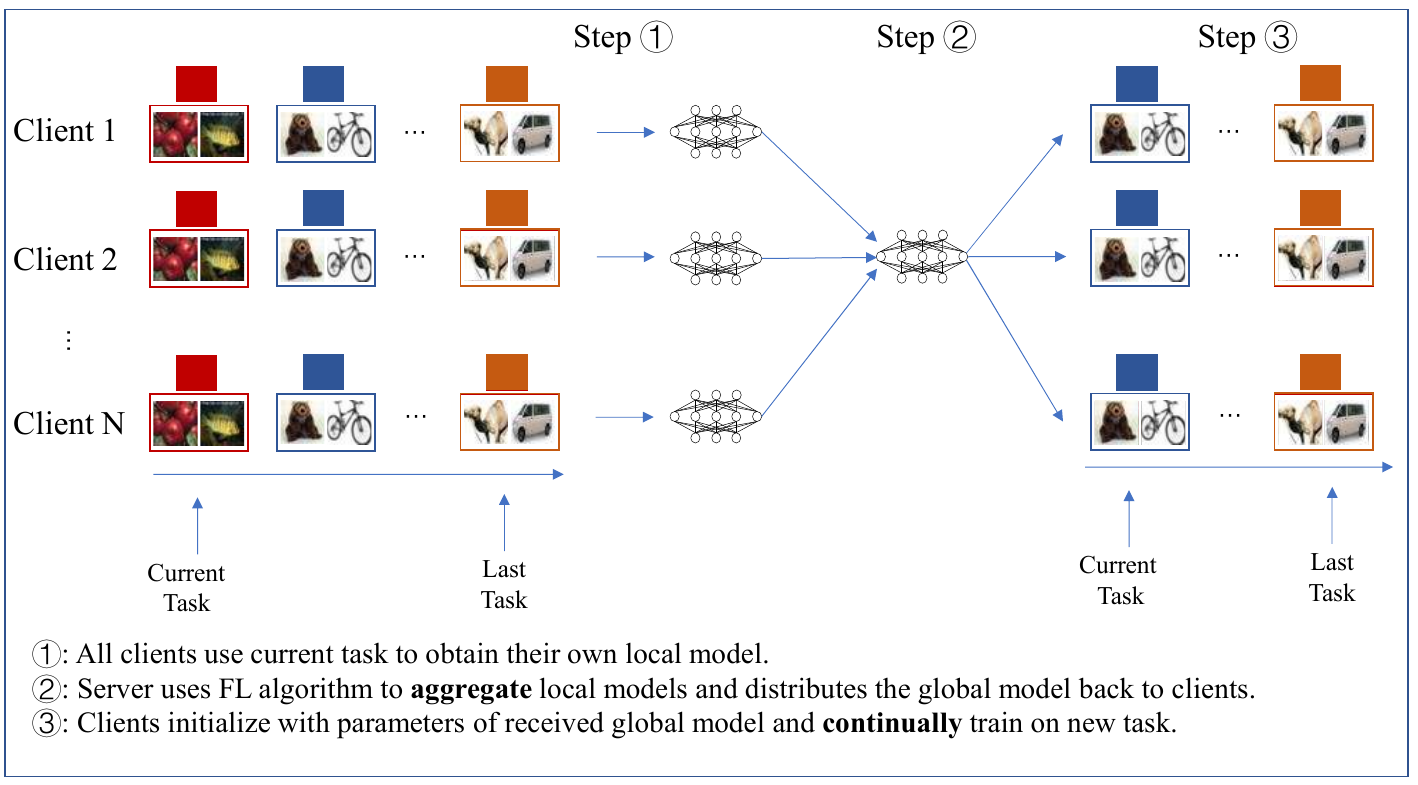}
  \caption{Illustration of \textbf{synchronous federated continual learning scenario}, in which all clients share a common task sequence. Different colors represent different tasks in a task sequence. Apparently, ``synchronous" means that clients will process the same task in a shared task order, and the progress of all clients is consistent. Synchronous FCL mainly focuses on \textbf{aggregation}.}
  \label{fig:fig1}
\end{figure*}

Based on whether the task sequence of the client is the same, FCL can be preliminarily divided into two types: synchronous FCL and asynchronous FCL.
Synchronous FCL refers that all clients share a common task sequence, as illustrated in Fig.~\ref{fig:fig1}.
In other words, all clients would learn one task and learn the next task together. 

Generally speaking, there are only a few approaches specifically designed to deal with this setting, as it requires that the similarity of distinct task sequences of clients should be as high as possible. According to the limitations we mentioned above, we prefer to call this scenario CFL rather than FCL. The setting is idealized since it has strict limitations on the task sequences of different clients. We give the definition of synchronous FCL:

\noindent\textbf{Definition 1.} A central server $S$ and $c$ clients, all the clients share the same task sequence $\mathcal{T} = \{\mathcal{T}_{1}, \mathcal{T}_{2}, \mathcal{T}_{3},$ $ \ldots, \mathcal{T}_{n} \}$ , where $n$ represents the number of tasks. Global model $\theta_G^i$ is obtained after aggregating local models $\{ \theta_1^{i-1}, \theta_{2}^{i-1}, \ldots, \theta_{c}^{i-1}\}$ uploaded by clients after training on task $i-1$. 

This scenario was originally described in Bui $et\, al.$ \cite{bui2018partitioned} in order to federated train Bayesian Neural Network and continually learn for Gaussian Process models.
Ma $et\, al.$ \cite{ijcai2022p0303.} firstly give a clear definition of Continual Federated Learning (CFL) to it. And they proposed a framework called CFeD, which employs knowledge distillation based on surrogate datasets to mitigate catastrophic forgetting both on the server-side and client-side. The most interesting point is that a part of clients are used for learning the new task and others are used for reviewing the old tasks in order to alleviate catastrophic forgetting.
Guo $et\, al.$ \cite{guo2021towards} present a unified framework, termed Continual Federated Learning, together with a novel client and time drift modeling approach, to capture complex FL scenarios involving time-evolving heterogeneous data. 

Due to its strict limitation on task sequences, synchronous FCL can not be further classified since it is up to one single client. If one of the clients runs a class-incremental task sequence, this whole FL system is class-incremental. However, we believe that the asynchronous one is worth more attention due to its practicality. 

\subsection{Asynchronous FCL}
\subsubsection{Federated Task-Incremental Learning}
Asynchronous FCL refers that clients will train their local models using distinct sequences of tasks and then send information to the server. And these sequences consist of tasks from a global task set in different orders. Subsequently, the knowledge from clients needs to be fused into one global model, then it will obtain the ability to perform well on all the tasks that have been seen by clients. Notice that no matter how different the order or number of tasks the clients process, in the view of the server, each client just processes one unique task. The definition of asynchronous FCL is as below:

\noindent\textbf{Definition 2.} Give $c$ clients $\{ C_{1}, C_2, \dots, C_c\}$ and one central server $S$. This system owns a global task set $T = \{T_1, T_2, \ldots, T_n\}$ where $n$ represents the total number of tasks. Each client follows its individual order $O_i, 1\leq i \leq c$ to process $T$, which means that at the same time, clients would handle different tasks. At time step $r$, client $C_i$ trains its local model $\theta_i^r$ on task $T_{O_i^r}$ based on the global model $\theta_G^{r-1}$, where $T_{O_i^r}$ represents the $r_{th}$ task in order $O_i$. Then, $\theta_i^r$ has learned knowledge of task $T_{O_i^r}$ and tries to upload it to the server. The server $S$ would fuse knowledge from client $C_i$ with other clients' knowledge and previous knowledge from time $0$ to time ${r-1}$. After fusion,  the global model $\theta_G^{r}$ is obtained and it has knowledge of all clients from time $0$ to time ${r}$. And $S$ then distributes $\theta_G^{r}$ back to clients and clients train it on the next task.

\begin{figure*}[htbp]
  \centering
  \includegraphics[width=1\textwidth,height=0.45\textwidth]{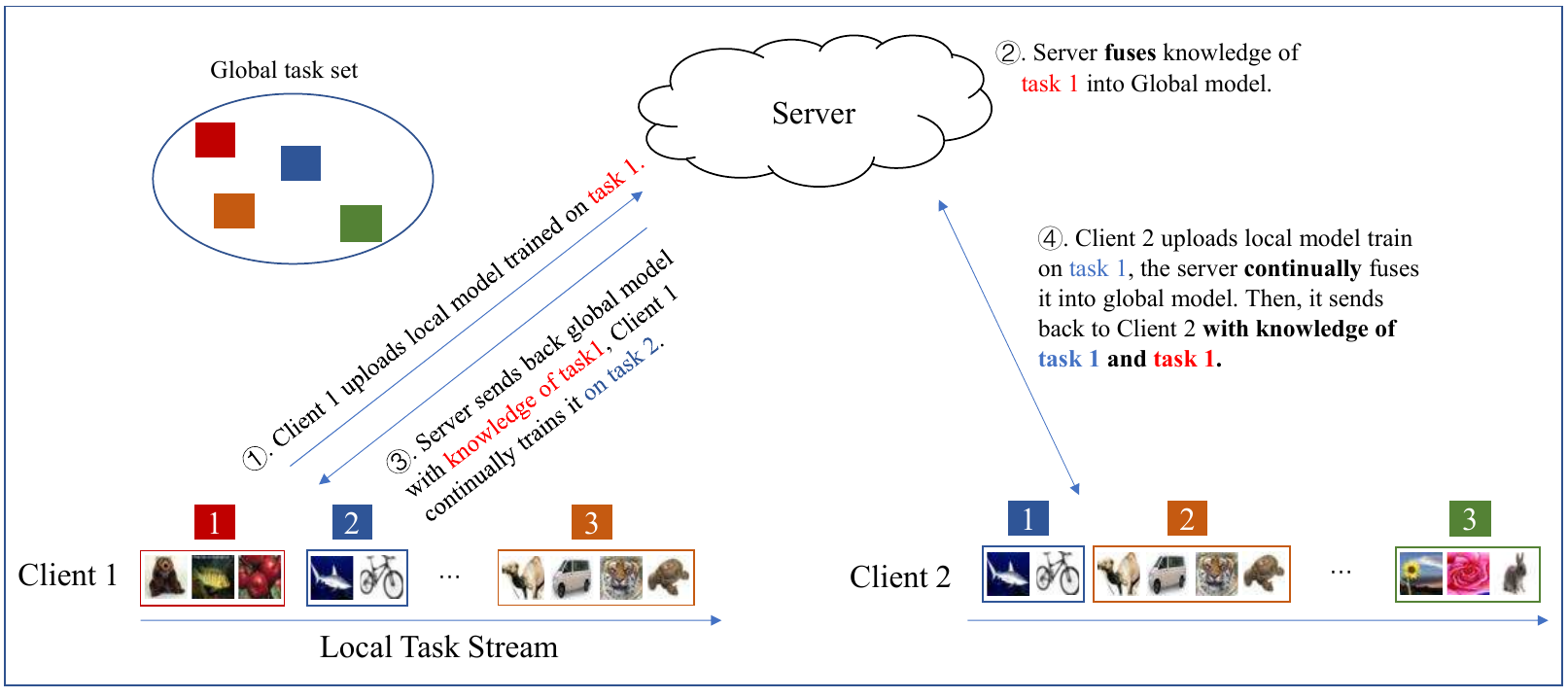}
  \title{}
  \caption{Illustration of Asynchronous FCL. In Asynchronous FCL, the process of fusing knowledge is gradual, which means that once a client has finished training on a task, the server timely fuses it into the global model without waiting for other clients. An ideal Asynchronous FCL can turn \textbf{aggregation} into \textbf{fusion}, allowing much more tolerance of asynchronous clients.}  \label{fig:fig2}
\end{figure*}

Here, we present an illustration of an ideal asynchronous FCL as shown in Fig.~\ref{fig:fig2}. Please note that we refer to this figure as an ``ideal'' asynchronous FCL because most existing FCL methods are primarily based on ``aggregation'' rather than ``fusion''. Therefore, they are more like a combination of asynchronous and synchronous approaches: handling different task sequences and then aggregating local models to obtain the global model rather than fusing knowledge into global models.

Apparently, asynchronous FCL is much more complex than the former one. As all clients share the same task sequence, there is no need to consider too much when aggregating models. However, in asynchronous FCL, the task sequences processed by clients are different, which poses greater challenges to overcome spatial catastrophic forgetting. On the one hand, knowledge needs to be layered, and on the other hand, different layers of knowledge need to be fused separately. But from another perspective, asynchronous FCL tends to be more practical because it is impossible to make the task sequence of all clients identical in reality.

The objective of asynchronous FCL is explicit: the global model needs to retain the knowledge of clients as much as possible when fusing local models. In some applications with higher real-time requirements, new local knowledge uploaded by clients also needs to be fused into global model continually. Besides, new local model which is initialized on global model in next round would be better if it retains some local parameters in order to prevent dropping performance on local tasks. 

To distinguish it from a specific kind of asynchronous FCL framework mentioned later, we will refer to the one described in Definition 2. as ``\textbf{Federated Task-Incremental Learning}".




\subsubsection{Federated Class-Incremental Learning. }
There's a special kind of asynchronous FCL called \textbf{Federated Class-Incremental learning}, which poses a greater challenge to heterogeneous knowledge fusion. 

Federated class-incremental learning involves an interesting setting where every client is expected to learn a growing set of classes and communicate knowledge of those classes efficiently with other clients, such that, after knowledge merging, the global model should be able to accurately discriminate between classes in the superset of classes observed by the set of clients. That is, the local model $\theta_i$ has the ability to identify class set $A_i$ on client $C_i$, and $\theta_j$ also can identify class set $A_j$. After fusing on a server, the global model would be able to identify $A_j \cup A_i$. Here, we give a form definition of federated class-incremental learning.

\noindent\textbf{Definition 3.} 
Consider $c$ clients, denoted as $\{C_1, C_2,$ $ \ldots, C_c\}$, and one central server, denoted as $S$. Each client $C_i, 1 \leq i \leq c$, has its unique task sequence $T_i$, which can differ significantly from one client to another. A set of public classes, denoted as $A_{public}$, is accessible to all clients, while each client $C_i$ has its private class set $A_i$. The primary objective of the local model $\theta_i$ is to incrementally learn to discriminate classes from the set ${A_i \cup A_{public}}$.

The task sequence of client $C_i$ is denoted as $T_i=\{T_i^1, T_i^2, \ldots, T_i^{n_i}\}$, where $n_i$ represents the total number of tasks on client $C_i$. The $k$-th task of $T_i$ contains $\left| A_i^k \right|$ classes, where $A_i = \{A_i^1\cup A_i^2\cup \ldots, \cup A_i^{n_i}\}$.

At time step $r$, the global model $\theta_g^{r-1}$ can distinguish $\left|A_g^{r-1}\right|$ classes. The server $S$ then distributes it back to clients. Client $C_i$ uses $\theta_g^{r-1}$ as an initial model to train on its $r$-th task $T_i^r$. The local model $\theta_i^r$ should perform well in classifying classes from the set ${A_g^{r-1} \cup A_i^r}$.

Finally, the server collects the local models from clients who participate in FCL and obtains a new global model $\theta_g^r$, which can identify classes from the set $A_g^{r} = \{A_g^{r-1} \cup \ A_1^r \cup A_2^r \cup \ldots, \cup A_c^r\}$.

To our best knowledge, this area is novel and only a few papers come out, as it includes the common real-world problem of incrementally learning new classes of objects. We believe that research in this area has enormous potential in the future according to the analysis of our experiments, so we will provide detailed descriptions of the papers mentioned later to inspire further research.

To better categorize the mentioned FCL frameworks, we provide Tab.~\ref{qqqwww} to help readers classify them more effectively.
\begin{table*}[htbp] 
\centering
\caption{How to distinguish the FCL framework?}
\label{qqqwww}
\vspace {-2.5mm}
\begin{tabular}{ccccc} 
\hline
\multicolumn{2}{c}{Framework}& {Same Task Sequence?}& {Global Task Set?} & {Private Classes on clients?}\\
\hline
\multicolumn{2}{c}{Synchronous FCL} &\cmark &\xmark &\xmark\\

\multirow{2}{*}{Asynchronous FCL} & Federated Task-Incremental Learning &\xmark&\cmark&\xmark\\
&Federated Class-Incremental Learning&\xmark&\xmark&\cmark\\

\hline 
\end{tabular}

\end{table*}

\section{Knowledge Fusion in FCL}\label{Sec5}
This section mainly discusses a diverse form of knowledge fusion in FCL. As discussed before, simply uploading the local model gradients to the server not only poses privacy risks but also hinders knowledge fusion. Recent research presented at NeurIPS 2019 \cite{zhu2019deep} demonstrated that in just a few iterations, a malicious attacker can fully extract the training data from gradients.
Yang $et\, al.$ \cite{10110919} recover images by using the approximate gradient in gradient leakage attack against Unbiased Sampling-Based secure aggregation.

Specifically, how to fuse the local knowledge uploaded by the clients on the server side (we call it \textit{collect and fuse} on the server), and how to fuse the knowledge distributed by the server with local knowledge on the client side (we call it \textit{receive and fuse} on clients), are the key issues in solving FCL problems. Recently, many researchers have noticed this and started to actively explore knowledge fusion forms other than the original gradient.  

\begin{figure*}[htbp]
	\centering
	\includegraphics[width=0.9\textwidth]{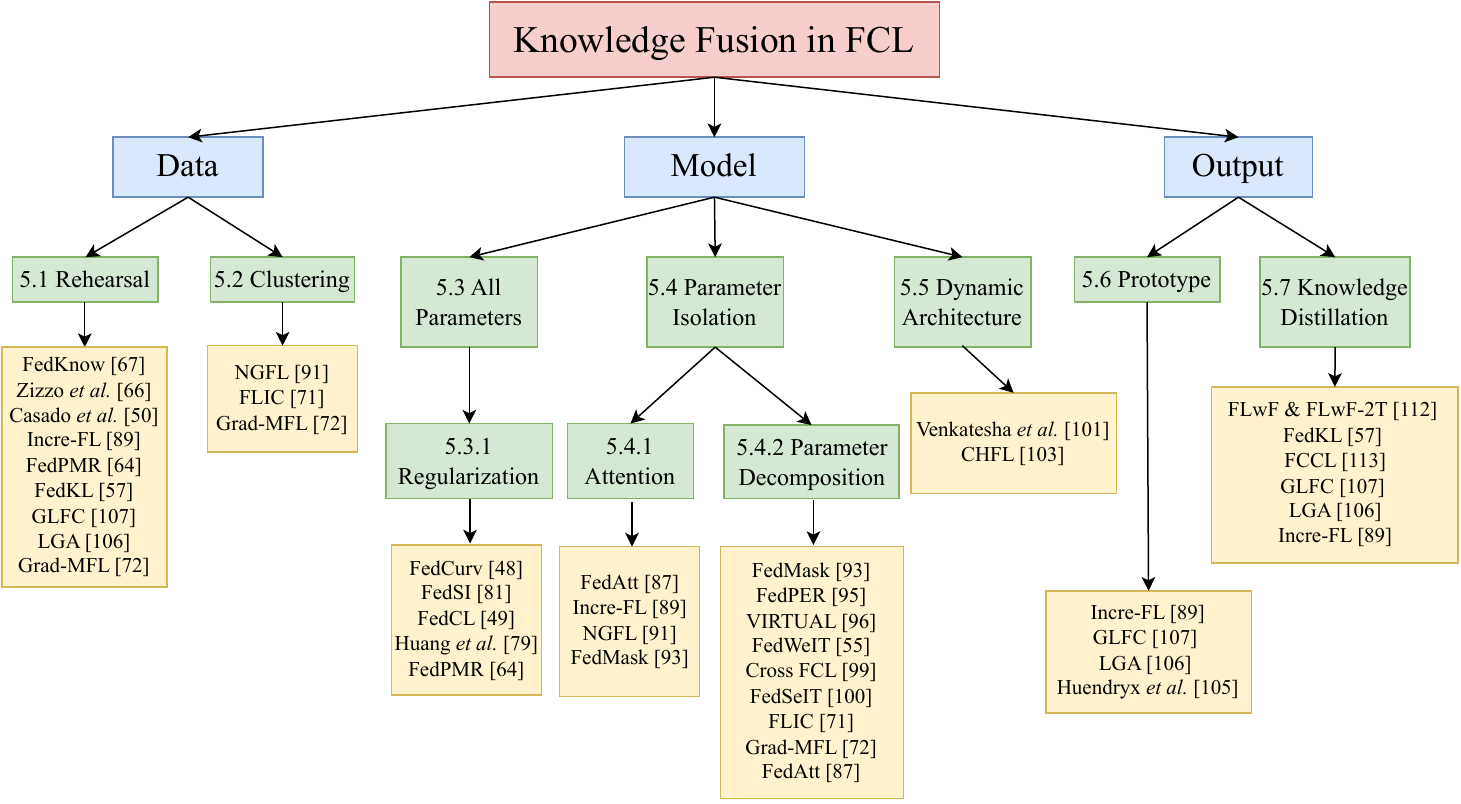}
    \caption{Knowledge Fusion in FCL.}
    \label{fig:structure}
\end{figure*}

Knowledge can be expressed in many forms. We find that in the FCL setting, local knowledge is hidden in three parts: data, models and outputs. We divide existing knowledge fusion methods into seven classes as shown in Fig.~\ref{fig:structure}. As FCL is an emerging field, some of the papers mentioned below may be of traditional FL, but their techniques for mitigating spatial catastrophic forgetting can be easily extended to FCL. According to the mechanism involved in knowledge fusion, we summarize existing methods with seven parts: \textit{Rehearsal} (\textbf{Sec.~\ref{sec5.1}}), \textit{Clustering} (\textbf{Sec.~\ref{sec5.2}}), \textit{All Parameters} (\textbf{Sec.~\ref{sec5.3}}), \textit{Parameter/Layer Isolation} (\textbf{Sec.~\ref{sec5.4}}), \textit{Dynamic Architecture} (\textbf{Sec.~\ref{sec5.5}}), \textit{Prototype} (\textbf{Sec.~\ref{sec5.6}}), \textit{Knowledge Distillation} (\textbf{Sec.~\ref{sec5.7}}).

\subsection{Rehearsal}\label{sec5.1}
Rehearsal-based approaches \cite{rebuffi2017icarl, shin2017continual} alleviate catastrophic forgetting by storing exemplars of previous data and replaying them when training on new data. Rehearsal-based approaches have been proven to achieve the best results in most CL settings\cite{van2019three}, since they ensure the memory of old knowledge will not degrade through time by keeping exemplars.

\textbf{FedPMR} (Federated Probability Memory Recall) \cite{wang2023federated} is a rehearsal-based FCL approach proposed by Wang $et\, al.$, which consolidates the memory of old probability experience and solves probability bias occurring in previous tasks with Probability Distribution Alignment (PDA) module. To be more specific, the PDA module maintains a replay buffer to hold a small number of exemplars for past tasks and the corresponding original probability outputs.

Casado $et\, al.$ \cite{casado2021concept} extend their research in \cite{casado2020federated}. They divide the replay buffer on each client into two parts: a short-term memory part and a long-term memory part. The former is processed to check whether a global drift occurs, and the latter is consisted of past concepts. 

In previous FCL methods, replay buffer is private for each client and exemplars will not be shared in order not to break the privacy protocol. However, due to the data heterogeneity among clients, the direct application of replay fails to achieve excellent performance. Zizzo $et\, al.$ \cite{zizzo2022federated}, therefore, creates a global replay buffer shared among all clients to mitigate forgetting more efficiently. Exemplars are processed with Laplace noise to avoid privacy leakage. They also consider a challenging setting of FCL, in which the clients dynamically participate in the FL system, and the fraction of participation varies in training rounds.  

Although a differential privacy mechanism can be employed to enhance security, rehearsal methods with shared buffers are prone to privacy breaches. Pseudo-rehearsal is thus introduced to FCL. The replayed data in this approach are generated to simulate the feature distribution of previous tasks. The advantages of this method are reduced memory requirement and enhanced privacy protection of old tasks.

\textbf{FedKNOW} \cite{luopan2022fedknow} acts in each client and extracts compact and transferable knowledge (instead of data) – the critical subset of model weights. When learning a new task, FedKNOW integrates it with the knowledge of its signature tasks, which are the new task’s most dissimilar tasks identified from local past tasks to prevent catastrophic forgetting, and the updated global model representing other clients’ current tasks in preventing negative knowledge transfer. By completing knowledge integration with polynomial time complexity, FedKNOW addresses the limitations of existing techniques by providing both high model accuracy and low communication overhead at the edge.

Unlike naive rehearsal or pseudo-rehearsal approaches, \textbf{GradMA} \cite{luo2023gradma} does not directly optimize on stored or generated exemplars but treats exemplars as inequality constraints and performs quadratic programming (QP) to correct the current gradient directions of global and local models in FCL setting iteratively. On the client side, GradMA utilizes various information, which includes the gradients of the local model in the previous round, the gradients of the global model, and the parameter differences between the local and global models in the current round, to adaptively correct the update direction of the local models. On the server side, GradMA utilizes the memorized accumulated gradients from all participants as constraints and performs QP to enhance the update direction of the global model.

\subsection{Clustering}\label{sec5.2}
Clustering is a technique used in unsupervised learning to group similar data points together based on certain similarities or patterns within the dataset. Some FCL methods cluster the tasks for each client or directly cluster clients according to their similarity of task flows to enhance the effectiveness of model aggregation. Clustering in FCL is typically for personalization, which aims to precisely assign greater importance to the specific information of individual clients in order to balance the generalized knowledge learned from the entire set of clients with their unique characteristics \cite{kulkarni2020survey}. There are two levels of personalization in FCL: client-level and group-level. Client-level personalization involves clients receiving the same global model from the server and adapting it locally to better fit their own tasks, as mentioned before. However, empirical evidence suggests that in practical scenarios, a single global model may not be able to accommodate the specific requirements of all clients \cite{sattler2019robust}. Group-level personalization emerges to address this issue, by clustering the clients according to their similarity distance of tasks, and jointly training multiple global models to better accommodate the clients with Non-IID datasets \cite{criado2022non}.

 We must take into account that the same input may produce different or even opposite outputs on different clients \cite{yu2020salvaging}. Therefore, it is necessary to consider adopting multiple global models. It is likely that this area of group-level personalization has not yet reached its full potential since it emerged a few years ago.

\cite{castellon2022federated} finds that there are groups of clients who have similar data distributions that can be clustered. It alleviates catastrophic forgetting by adding a clustering step to separate into groups according to the similarity of tasks among clients. The resulting global model after clustering will be more effective, as it excludes knowledge uploaded by clients with low similarity.

Gradient Memory-based Federated Learning (\textbf{GradMFL}) \cite{tong2022gradmfl} clusters the entire Non-IID data among clients into several IID data groups based on their similarity. Additionally, it introduces a collaborative training strategy for performing a series of tasks in hierarchical clusters, which incorporates gradient memory to mitigate the problem of catastrophic forgetting during the transfer of hierarchical knowledge.
 
It is undeniable that the personalization strategy provides a new idea for promoting knowledge integration: The knowledge learned by each client can be divided into two parts: local knowledge and global knowledge. In the aggregation stage, only the global knowledge needs to be aggregated, which reduces communication costs and preserves privacy. If the similarity between the tasks of the clients is too low or even completely opposite, we can group users and obtain more than one global model. Although this increases computational costs, it can better aggregate similar knowledge and protect the system from a series of attacks \cite{lyu2020threats,bagdasaryan2020backdoor}.

\subsection{All Gradients/Parameters}\label{sec5.3}

FedAvg \cite{Communication-efficientlearningofdeepnetworksfromdecentralizeddata} brings out the concept of FL and proposes a simple weighted averaging approach to aggregate local model updates based on the amount of data. Most of the early works on FL \cite{zhong2021p,xia2021auto,collins2022fedavg} were simply making minor improvements to FedAvg, following its parameter transfer and aggregation-based paradigm. Even if the original data is not directly shared, the model parameters have the potential to reveal sensitive information about the training data \cite{shokri2017membership,nasr2019comprehensive}. Furthermore, this form fails to alleviate catastrophic forgetting of previously learned knowledge during the fusion process and does not facilitate the knowledge integration between local and global models, as it solely relies on weighted averaging of parameters. When confronted with class-incremental challenges, as shown in Sec.~\ref{Sec3}, the performance would significantly decrease.

\subsubsection{Regularization}

Regularization, a continual learning approach,  aims to mitigate forgetting by adding a regularization term to the loss function to constrain the direction of parameter optimization, making past knowledge less likely to be forgotten\cite{kirkpatrick2017overcoming}. Due to the intrinsic similarity between spatial and temporal forgetting in FCL, either of them can be alleviated with regularization-based approaches.

Although this technique still uses all parameters to aggregate models, it adds regularization loss terms to the loss function to constrain the direction of parameter optimization, making past knowledge less likely to be forgotten. In FCL, preserving local knowledge from each client during the global model aggregation process requires the inclusion of corresponding regularization terms into the global model. Similarly, if the goal is to retain previously learned knowledge, adding regularization terms would be sufficient.

FedCurv\cite{shoham2019overcoming} firstly adopts this approach in FCL. It follows the method in EWC \cite{kirkpatrick2017overcoming},  utilizing the diagonal of the Fisher information matrix to evaluate the importance of network weights for past tasks and adding a penalty term to the client-side loss function to ensure that local models can converge to a shared optimal solution.  However, FedCurv only adopts regularization to handle statistical heterogeneity, leaving temporal catastrophic forgetting unsolved.

\cite{huang2022continual} applies synaptic intelligence (SI) \cite{zenke2017continual}, which is applied to preserve important model weights for training one center after another, to FCL settings in order to identify automatically brain metastasis (BM) as an exemplary case of multi-center collaboration.

\cite{zhang2023communication} proposes a novel federated continual learning method called FedSI, also integrating SI with FL. The authors design an optimization loss function that leverages knowledge from multiple sources. This function guides the local training of all local models over a common parameter space by combining two loss terms. The first term, which is the cross-entropy loss, is calculated with the local training dataset to learn its own data. The second term, which is the proposed CL loss, helps to alleviate the weight divergences of local models by controlling the difference between the local model and the other local models. This knowledge fusion strategy compels the combined global model to move closer to the global optimal solution.

In \cite{yao2020continual}, Yao $et\, al.$ also adopt EWC and propose federated learning with local continual training (\textbf{FedCL}) strategy to alleviate the weight divergence and continually integrate the knowledge on different local clients into the global model, which ensures a better generalization ability for the global model. In contrast to FedCurv, which estimates the importance weight of model parameters on clients and exchanges them, leading to at least twice the extra communication costs, FedCL adopts a different approach. Specifically, FedCL estimates the importance weights on the server using a proxy dataset and then distributes these weights to the clients. Then clients subsequently utilized to restrict the local training. This approach enables the federated model to acquire knowledge from clients while preserving its original performance.

Regularization methods are more efficient when dealing with similar tasks since their main purpose is to limit the directions of parameter updates and achieve a shared solution. These approaches can not learn to discriminate classes from different tasks in the class-incremental scenario \cite{lesort2019regularization}. Therefore, to enhance knowledge fusion in both space and time dimensions, FCL requires more complicated strategies.

\subsection{Parameter Isolation}\label{sec5.4}

Uploading all parameters or gradients may make local models expose to reconstruction attacks or model inversion attacks when the server is honest but curious\cite{geiping2020inverting}. As a result, in some methods, local parameters are partitioned according to special rules, and only a portion of them are uploaded. Intuitively, this technique can be combined with parameter isolation in CL, which dedicates different parts of parameters to each task, to jointly avoid privacy leakage and overcome forgetting.

The parameter isolation approach in FCL can be split into two steps: local isolation and global knowledge fusion. Local isolation can be achieved through attention mechanism or parameter decomposition. Attention works by focusing on specific parts or features of input data that are deemed important for a particular task. Parameter decomposition is a CL approach proposed in \cite{yoon2019scalable}, which treats network weights as a sum of task-shared and sparse task-adaptive parameters and decomposes them to prevent forgetting and order sensitivity. Global knowledge fusion refers to the uploading and aggregation of the decomposed and selected information. 

Here, we provide a detailed description of how the attention mechanism and parameter decomposition work in FCL.

\subsubsection{Attention}

The attention mechanism, initially popularized in the field of computer vision, is a vector that directs the focus of perception. \cite{mnih2014recurrent} incorporates it into recurrent neural network models for image classification. Subsequently, attention played an important role in the field of natural language processing (NLP) \cite{vaswani2017attention}. \cite{luong2015effective} expands the application of attention-based RNNs by proposing two novel mechanisms: global attention and local attention. Hence, it can easily be adapted to federated settings. 

The work of \cite{ji2019learning} pioneers the use of attention mechanism in the aggregation of multiple distributed models. To achieve knowledge fusion during server optimization, they introduce a layer-wise soft attention mechanism to capture the ``attention" among local models' parameters. This approach enables the automatic selection of the weights for different client models, enabling them to minimize the expected distance between the server model and client models. \cite{estiri2021attentive} proposes \textbf{FedAtt} to tackle concept drift in distributed 5G networks and achieves excellent results. FedAtt takes the server as query and clients as keys, computes layer-wise attention scores, and minimizes the weighted distance between global and local models, which enables the FL system to adaptively evaluate and balance the contribution of each local model. Notice that unlike the original attention mechanism, which manipulates the input data flow, on the contrary, attentive aggregation in FCL usually works on model weights to control the intermediate results in the entire training procedure.

Moreover, the attention mechanism is also used in Class-Incremental FCL. 
The proposed approach in \cite{hu2022federated} commences with the random selection of a balanced set of samples from each client, establishing a balanced pre-training dataset. The federated averaging technique is then applied to train the model, resulting in an initial global model on the server. Subsequently, the traditional FL framework is augmented with the iCaRL strategy \cite{rebuffi2017icarl}. Notably, a dual attention mechanism is integrated into this framework.
The authors use a Channel Attention Neural Network model that integrates the SE module \cite{hu2018squeeze} into the Graph Convolutional Neural (GCN) network as the FL local model. This model allows for the identification of the significance of features for each client's overall samples during training and effectively reduces noise interference. The authors design a federated aggregation algorithm based on the feature attention mechanism to assign appropriate attention weights to each local model for the global model. These weights correspond to the model parameters of each layer of the neural network and act as aggregation coefficients to enhance the global model's ability to capture essential information and features.

\cite{li2022new} further uses the attention mechanism to alleviate both local and global forgetting. Their preliminary experiment illustrates the existence and impact of local forgetting. On a single client, if there is no intervention, the previously learned knowledge will deteriorate quickly when the new tasks do not overlap with the old ones. This problem is not limited to the client side; it also affects the server side since the heterogeneity of the client-side data and aggregation can cause some knowledge to be forgotten. Moreover, the dynamic nature of tasks on the client side exacerbates this forgetting phenomenon.
To address this, they propose a solution: Self-Attention (for only the current task) and Total Attention (for all observed classes).

Hard attention, also called ``mask", is also applied to FCL. \cite{serra2018overcoming} first uses hard attention in the Task-IL scenario of CL. The idea of partial initialization is firstly applied to FL by \cite{zhu2021initialize}. The authors have identified that a significant amount of local knowledge is lost when using the global model to initialize local models in FL. In order to address this issue, they have implemented a partial initialization approach that aims to preserve some local parameters during the initialization process. \textbf{FedMask} is based on a new encryption technology called ``mask", which allows participants to share only a small part of their model (that is, mask) rather than complete model parameters. In each federated learning iteration, participants will use their own data to locally update the shared mask, and then send the updated mask back to the server. The server will aggregate the masks of all participants and apply them to the initialization of the global model in order to fuse global knowledge and retain local knowledge.

\subsubsection{Parameter Decomposition}
Parameter decomposition approach is based on Yosinski $et\, al.$'s research \cite{yosinski2014transferable}, which concludes that the feature extraction component (shallow layers) of a neural network is comparatively generic because they do not acquire task-specific knowledge. Intuitively, it can be applied in FL by partitioning parameters into task-specific and global parts to separate the private knowledge of local tasks and the common knowledge shared among all clients.

In \textbf{FedPER} \cite{arivazhagan2019federated}, Clients train a Deep Neural Network (DNN) where the last few layers are not shared, and each client trains them separately while sharing the earlier layers as the global model. The last layers act as the personalization model, enabling different participants to obtain distinct results for comparable inputs. This approach not only facilitates the integration of global knowledge but also preserves local knowledge to some extent.

\cite{corinzia2019variational} introduces  \textbf{VIRTUAL} (VarIational fedeRaTed mUltitAsk Learning) to address multi-task learning in FL settings. Every client has a task-specific model that benefits from the server model in a transfer learning fashion with lateral connections. A part of the parameters are shared between all clients, and another part is private and tuned separately. The server maintains a posterior distribution that represents the plausibility of the shared parameters. 

Federated Weighted Inter-client Transfer (\textbf{FedWeIT}), a novel framework proposed by Yoon $et\, al.$ \cite{yoon2021federated}, addresses spatial-temporal forgetting in a different way. It decomposes network weights into two parts: global federated parameters and sparse task-specific parameters. It is derived from Additive Parameter Decomposition (APD) proposed in \cite{yoon2019scalable}, which alleviates forgetting by separating weights into task-shared and sparse task adaptive parameters, and keeping task adaptive parameters for previous tasks unaffected when training on new tasks. By taking a weighted combination of other clients' task-specific parameters, each client acquires selective knowledge from them. The sparse parameter selection technique not only reduces interference between clients but also facilitates efficient communication. FedWeIT effectively mitigates the interference between incompatible tasks, while facilitating positive knowledge fusion among clients during the learning process.

The idea of parameter decomposition is also adapted in mobile edge computing (\textbf{MEC}) systems \cite{lopez2021digital,8961984}. \textbf{Cross-edge FCL} \cite{zhang2022cross} separates the knowledge of the local model into two kinds of parameters, in which the base parameters learn the general knowledge between different tasks, and the task-specific parameters learn the personalized knowledge of the current local task. And the cross-edge strategies deal with the relationship between the local old task of the dynamic device and the global task of the new FL system.

\cite{chaudhary2022federated} is the first one who applies FCL to NLP. The authors propose a novel framework, Federated Selective Inter-client Transfer (\textbf{FedSeIT}), which efficiently selects the relevant task-adaptive parameters from the historical tasks of other clients assessing domain overlap at the global server using encoded data representations while preserving privacy. The FedSeIT model decomposes the model parameters of each client into three distinct sets of parameters: (1) dense local base parameters, which cover and accumulate task-agnostic knowledge on the client's private task sequence, (2) sparse task-specific adaptive parameters, which capture task-specific knowledge for each task across different clients, and (3) sparse mask parameters, which allow the client model to selectively leverage global knowledge.

In summary, this method divides parameters into relevant and unique parts. The knowledge fusion during the aggregation stage only involves activating the relevant parameters while keeping the client-specific part unchanged.

\subsection{Dynamic Architecture}\label{sec5.5}
While the parameter decomposition approach requires a fixed capacity of the model, dynamic architecture approach allows network size or connection paradigm to change if needed. 

\cite{venkatesha2022addressing} designs a structured FCL framework based on NetTailor \cite{morgado2019nettailor} , which involves a common backbone network trained on a large dataset such as ImageNet and the task-specific blocks interspersed between the layers. It allows the direct creation of links between the backbone network and specific layers, bypassing others.

Mori $et\, al.$ \cite{mori2022continual} find that most Horizontal Federated Learning (HFL) works such as \textbf{FedProx} \cite{li2020federated} only use common features space and leave client-specific features unutilized. Their framework Continual Horizontal Federated Learning (\textbf{CHFL}) splits the network into two columns corresponding to common features and unique features, respectively. The first column is trained jointly using common features via vanilla HFL, while the second column is trained locally using unique features and lateral connections that leverage the knowledge of the first column, without disrupting the federated training process.

In traditional continual learning, the parameter isolation approach can be categorized into fixed network structure and dynamic network methods. In the fixed network method, only the relevant parameters are activated for each task, without modifying the network structure \cite{mallya2018packnet,fernando2017pathnet}. In the dynamic network method, the network structure is modified to add new parameters for new tasks while keeping the old parameters unchanged \cite{rusu2016progressive,aljundi2017expert}. However, the situation is different in FCL. The dynamic architecture method mainly focuses on changing the network connections rather than continually increasing the number of parameters, while the parameter separation method mainly focuses on separating client parameters from global parameters. Furthermore, the parameter separation approach is widely used in FCL, and there are many related studies on its application. For the above reasons, we classify the parameter separation approach and dynamic architecture approach as two separate categories, rather than following the way of grouping them under the broader category of parameter isolation.

\subsection{Prototype}\label{sec5.6}
The prototypical network is a widely used approach in few-shot learning \cite{snell2017prototypical}, which learns a representation space where samples from the same class are clustered together. Given a few examples of a new class, the prototypical network can quickly learn to classify new samples into the corresponding class based on the similarity to the prototypes of each class. \cite{hendryx2021federated, dong2023no, dong2022federated} attempt to  represent the knowledge of clients in FCL using this approach. Specifically, they try to compress concepts into relatively small vectors in a common embedding space. Each class has its own prototype. During the aggregation stage, clients upload their own prototypes to the server, and then the server combines and extends the categories of the same class. Subsequently, the merged prototypes are distributed to the clients, completing the knowledge exchange between local and global. Moreover, for privacy preservation, prototypes can be perturbed while transmitting. And it requires only a small amount of communication and reduces memory storage costs, leading to a more efficient federated system. Here are detailed introductions of these works. Notice that the above three papers are all about class-incremental FCL.

The first attempt to systematically learn a global class-incremental model in FL settings is \cite{dong2022federated}. 
Dong $et\, al.$ pave a new approach to tackle both local and global catastrophic forgetting, known as the Global-Local Forgetting Compensation (\textbf{GLFC}) model. In this model, they propose a class-aware gradient compensation loss to mitigate the local forgetting caused by class imbalances at the local level, which balances the forgetting of different old classes. They also introduce a class-semantic relation distillation loss to maintain consistent inter-class relations across different incremental tasks. To address global catastrophic forgetting, they design a proxy server that selects the best old global model for class-semantic relation distillation at the local level. At last, to ensure privacy preservation, the proxy server adopts a prototype gradient-based communication mechanism to gather perturbed prototype samples of new classes from local clients. These samples are then utilized to evaluate the performance of the global model and select the most optimal one. 

However, Li $et\, al.$ \cite{li2022new} argue that the global-local compensation in \cite{dong2022federated}, is idealistically assumed and ignores some key issues. The first neglected issue is task overlap. True incremental tasks and pseudo-incremental tasks should be distinguished to avoid unnecessary overhead. There are three scenarios for client-side incremental tasks: full-covered, semi-covered and not covered. If the present classes of a specific client are fully covered by its previous classes, the client can be trained directly. The authors solve this problem by utilizing a double-ended task alignment table. Similar to the routing table, it exists on both the server and the client side. The table on the client side records all task classes covered up to the current task and is uploaded to the server along with the local model after local training. The table on the server side then participates in the knowledge fusion of local models. The second neglected issue is the aggregation of heterogeneous local models. Due to the intrinsic nature of continual learning, different clients have dimensionally identical feature extractors and structurally heterogeneous classifiers (the output layers). The authors put forward Pre-Alignment (PreA) and Post-Alignment (PostA) strategies to solve this problem. PreA method means the output layers of clients are pre-defined by the server before each round of federated training. PostA method means the server adjusts and aggregates the output layers according to the alignment table submitted by the client. Another approach to perform the aggregation of heterogeneous local models is to utilize partial masks for the specific knowledge of each client and a total fusion mask for common knowledge among clients \cite{li2021fedmask}.

Then Dong $et\, al.$ update their work. The different forgetting speeds of old classes resulting from the data heterogeneity among clients can be significantly alleviated by \textbf{LGA} (Local-Global Anti-forgetting) \cite{dong2023no}. Specifically, LGA surmounts both forgetting by a category-balanced gradient-adaptive compensation loss and ensures intrinsic category relations consistency within different incremental tasks with a category gradient-induced semantic distillation loss. Besides, perturbed prototype images of new classes are transmitted from local clients to the proxy server, reconstructed by the proxy server and processed via self-supervised prototype augmentation to pick the best old global model for semantic distillation. Global forgetting is thus overcome by improving the distillation gain of the category gradient-induced semantic distillation loss at the local side by providing the best global model from a global perspective.

Hendryx $et\, al.$ \cite{hendryx2021federated} call the class-incremental federated learning the federated reconnaissance. In traditional federated learning, the set of classes learned by each client is static. Federated reconnaissance, instead, requires that each client individually learn a growing set of classes and effectively communicate knowledge of previously observed and incoming classes with other clients. \cite{hendryx2021federated} proposes prototypical networks to address federated reconnaissance, which compresses concepts into relatively small vectors known as prototypes, enabling efficient communication. More important is that it enables fast knowledge transfer as it does not rely on gradient. Instead, first, a base model is trained on a set of base classes to provide the foundation for basic classification. The model is then deployed to clients, who are trained on instances of new or previously seen classes using local supervision. After each client has learned new classes, it is evaluated on the combined set of previous and new classes. The clients then share their knowledge of the new classes with a central server, which merges the knowledge and distributes it back to the clients. At this point, the server and clients have prototypes for each class, which can be used for classification. The most interesting point about this paper is that it can restore and load knowledge extracted from other tasks, as Knowledge Base used in CL \cite{yang2022three, schwarz2018progress}.

\subsection{Knowledge Distillation}\label{sec5.7}
Knowledge distillation is a popular technique to fuse knowledge, the core idea of which is to distill the knowledge contained in an already trained teacher model into a student model \cite{hinton2015distilling}. Basically, the teacher model teaches students to construct the same mapping relationship by outputting the label $Y$ corresponding to sample $X$, just like humans. This technique is widely used in FCL \cite{wei2022knowledge,dong2022federated,usmanova2021distillation2}. 

\cite{dong2022federated} designs a class-semantic relation distillation loss to maintain consistent inter-class relations across tasks during the extraction/distillation process. 

Inspired by Learning without Forgetting (LwF) \cite{li2017learning}, Knowledge Distillation is implemented in class-incremental FCL \cite{usmanova2021distillation2}. Regarding the past model of the client as the first teacher model and the current client model as the first student model, the final loss for each client consists of a classification loss and a distillation loss, the latter is computed with the current model and previous model of the same client. Then, regard the global model as the second teacher model to make use of the server's ability to maintain a comprehensive knowledge base of all clients. The first teacher (the past model of a client) can improve the performance of a student on specific tasks, helping it maintain proficiency on previously learned tasks. On the other hand, the second teacher (the server) enhances the general features of a client model by transferring knowledge from all other clients and mitigating the risk of overfitting on new tasks.

Moreover, \textbf{FedKL} in \cite{wei2022knowledge} decouples the training objective into a classification and a knowledge maintaining term. In the classification term, the local training is supervised by the original label of the data with a softmax-cross-entropy loss applied. In the knowledge maintaining term, the locally unavailable classes are supervised by the distillation of the global model with logic regression loss. 

\cite{huang2022learn} proposes \textbf{FCCL} (Federated Cross-Correlation and Continual Learning), which  utilizes knowledge distillation in local updating, providing inter and intra-domain information without leaking privacy. By utilizing unlabeled public data and implementing self-supervised learning techniques, heterogeneous models attain a generalizable representation while ensuring efficient communication.

\subsection{Summary}
It is worth noticing that some of the papers mentioned above may use more than a single technique to realize knowledge fusion. The following Tab.~\ref{tab:table2} provides a detailed description of the methods employed in each model.
 \begin{table*}[htbp]
\centering
\caption{A Summary of Knowledge Fusion Methods Used in Mentioned Paper.}
\label{tab:table2}
\resizebox{\textwidth}{!}{
\begin{tabular}{ccccccccc}
\hline
    & Regularization & Attention & Parameter Isolation & Clustering & Dynamic Structure & Prototype & Knowledge Distillation & Rehearsal \\ 
\hline
\multicolumn{1}{l}{FedAvg \cite{Communication-efficientlearningofdeepnetworksfromdecentralizeddata}}  &  \\

\multicolumn{1}{l}{FedCurv \cite{shoham2019overcoming}}  & \cmark &    \\
\multicolumn{1}{l}{FedSI \cite{zhang2023communication}}  &  \cmark& \\
\multicolumn{1}{l}{FedCL \cite{yao2020continual}} &  \cmark  &       \\ 
\multicolumn{1}{l}{Huang $et\, al.$ \cite{huang2022continual}} & \cmark  & \\
\multicolumn{1}{l}{FedAtt  \cite{ji2019learning}} &  &\cmark & \cmark & \\
\multicolumn{1}{l}{Incre-FL  \cite{hu2022federated}} &  &\cmark & &&&\cmark &\cmark&\cmark\\
\multicolumn{1}{l}{NGFL  \cite{li2022new}} &  &\cmark && \cmark & \\
\multicolumn{1}{l}{FedMask  \cite{zhu2021initialize}} &  &\cmark & \cmark & \\
\multicolumn{1}{l}{FedPER  \cite{arivazhagan2019federated}} &  & & \cmark & \\
\multicolumn{1}{l}{VIRTUAL  \cite{corinzia2019variational}} &  & & \cmark & \\
\multicolumn{1}{l}{FedWeIT   \cite{yoon2021federated}} &  & & \cmark & \\
\multicolumn{1}{l}{Cross FCL  \cite{zhang2022cross}} &  & & \cmark & \\
\multicolumn{1}{l}{FedSeIT  \cite{chaudhary2022federated}} &  & & \cmark & \\
\multicolumn{1}{l}{FLIC  \cite{castellon2022federated}} & & & & \cmark & \\
\multicolumn{1}{l}{Grad-MFL  \cite{tong2022gradmfl}} &  & & &\cmark & &&&\cmark \\
\multicolumn{1}{l}{Venkatesha $et\, al.$  \cite{venkatesha2022addressing}} & && & & \cmark & \\
\multicolumn{1}{l}{CHFL  \cite{mori2022continual}} & && & & \cmark & \\
\multicolumn{1}{l}{GLFC  \cite{dong2022federated}} & &&& & & \cmark &\cmark &\cmark \\
\multicolumn{1}{l}{LGA  \cite{dong2023no}} &  &&&& & \cmark & \cmark &\cmark\\
\multicolumn{1}{l}{Huendryx $et\, al.$  \cite{hendryx2021federated}} &  &&&& & \cmark & \\
\multicolumn{1}{l}{FLwF \& FLwF-2T  \cite{usmanova2021distillation2}} &  &&&&& & \cmark & \\
\multicolumn{1}{l}{FedKL \cite{wei2022knowledge}} &  &&&&& & \cmark &  \cmark \\
\multicolumn{1}{l}{FCCL  \cite{huang2022learn}} &  &&&&& & \cmark & \\
\multicolumn{1}{l}{FedPMR  \cite{wang2023federated}} &\cmark&&  &&&& & \cmark  \\
\multicolumn{1}{l}{Casado $et\, al.$  \cite{casado2021concept}} &  &&&&& && \cmark \\
\multicolumn{1}{l}{Zizzo $et\, al.$   \cite{zizzo2022federated}} &  &&&&& && \cmark \\
\multicolumn{1}{l}{FedKnow  \cite{luopan2022fedknow}} &  &&&&& && \cmark \\

\hline
\end{tabular}}
\end{table*}

In summary, to overcome the problem of spatial-temporal catastrophic forgetting in FCL, knowledge fusion is necessary. Knowledge can be represented in various forms: data, gradients, parameters, models and so on. How to fuse knowledge, especially fuse relevant knowledge, is the key to address challenges. Regularization realizes the common optimal solution by constraining the direction of parameter update, but if there is a huge difference between the two tasks, or even the opposite, then the performance will be even worse. The method of parameter isolation divides the parameters of a model into different parts based on specific rules, fusing relevant knowledge together without fusing irrelevant ones, which can further improve the generalization ability of the global model. Based on this idea, clustering divides clients into different groups through similarity, generating more than one global model, and avoiding the possibility of integrating irrelevant knowledge.

By utilizing specific rules to divide the parameters of a model, the method of parameter isolation only fuses relevant knowledge, ultimately enhancing the generalization ability of the global model. This idea is further developed by clustering, which groups clients based on similarity to generate more than one global model and prevent the integration of irrelevant knowledge.

The dynamic network structure method allows for dynamic adjustment of the network structure to accommodate more knowledge, but this puts higher demands on the design of the network structure. The knowledge distillation method conveys knowledge through the input and corresponding output of the models. The student model adjusts its output according to the guidance of the teacher model, making its output similar to the output of the teacher model, thus achieving knowledge transfer. Similarly, if encountering unrelated teacher models, its output may shift, resulting in the inability to complete tasks.

Both prototype and rehearsal methods express knowledge through data. Replay past knowledge to learn from it, but in FCL settings, a simple replay may cause privacy breaches. Moreover, whether replaying data from other clients is useful for oneself is also a question. And prototype networks achieve knowledge sampling by compressing a type of data into prototypes in a sample space. If we want to use this method in FCL, then we must ensure the consistency of the sample space.

Replay method mitigates forgetting by revisiting past knowledge, but in FCL setting, such a simple replay can cause privacy breaches. Additionally, it is unclear whether using data from other clients for replay is beneficial to one's own local training. In contrast, prototype networks use prototypes in a sample space to compress data and obtain knowledge samples. However, if we want to apply this method in FCL, it is necessary to ensure the consistency of the sample space.

\section{Future work and Discussion}\label{Sec6}
In this survey, we explore the origins of FCL and discuss why federated learning and continual learning should be combined. Then we firstly define a fundamental and main problem in federated continual learning called ``spatial-temporal catastrophic forgetting" and experiments have been conducted to verify the horrible impact on the performance of the global model, using FedAvg. 

In Sec.~\ref{Sec4}, we summarize two generic frameworks, namely synchronous FCL and asynchronous FCL. The focus of synchronous FCL lies in the \textbf{aggregation of local models}, which is a primary stage of FCL. Furthermore, asynchronous FCL tries to turn \textbf{aggregation based} into \textbf{knowledge fusion based}. We believe that knowledge-based FCL is the most promising future direction, although now existing methods are like a mixture of asynchronous and synchronous.

In Sec.~\ref{Sec5}, we classify the existing knowledge fusion technologies of Federated learning according to the expression forms of knowledge in different stages of model training. 

The following issues are worthy of exploring and we believe they could form future avenues of research.

\textbf{Trustworthy FCL}: One of the biggest challenges in promoting FL techniques is to strike a balance between privacy protection and model utility, which is also confronted by FCL. Generally, the enhancement of privacy comes at the cost of decreased model performance. To address this issue, Yang proposes Trustworthy Federated Learning to incorporate secure and reliable mechanisms into distributed federated modeling. \cite{zhang2022trading} first reveals the incompatibility of privacy and utility in FL, set it as a multi-objective optimization problem, and applies the no-free lunch theorem in FL \cite{zhang2022no} to three privacy-preserving mechanisms, namely differential privacy, secure multi-party computation and homomorphic encryption. In addition to privacy and utility, an FL method should also consider the efficiency of algorithm, robustness of outliers, fairness of sampling, incentive mechanism and interpretability of the whole model \cite{zhang2023survey}. \cite{kang2023optimizing} designs measurements of privacy leakage, utility loss and training cost for various privacy protection approaches, and finds Pareto optimal solutions to all considered objectives. The conflicts of objectives, especially the contradiction between utility and privacy, also exist in FCL. And the dilemma between stability and plasticity makes the problem more complicated and challenging. A possible solution is to seek the Pareto Front of multi-objective optimization and determine the equilibrium point based on the specific requirements of users, which is worth researching.

\textbf{Delayed Decision-Making: }According to the intrinsic nature of deep learning, the performance of the model is generally weak at the beginning and gradually strengthens during the process of learning consecutively coming samples \cite{yao2009three}. Obviously, if uncertain samples are subjected to delayed decision-making, the average accuracy can be significantly improved, which is the underlying assumption of three-way decision (3WD) \cite{yang2017unified,yang2019sequential}. 3WD divides the decision space into three mutually disjoint regions, respectively signifying positive, negative and uncertain, the decision-making of uncertain samples is delayed until the learner has sufficient capacity of predicting at a high level of confidence \cite{liu2020three}. Some scholars have already incorporated 3WD into CL, taking advantage of the forward transfer of knowledge in CL \cite{li2022incremental,yang2022three} to improve the performance of the model. However, there is not yet any study incorporating 3WD into FCL. 

\textbf{Rational Collaboration: }Making all clients collaborate does not necessarily result in the best performance. FL is profitable only when the benefits outweigh the costs. Cui $et\, al.$ \cite{cui2021collaboration} advances a rational collaboration method called "collaboration equilibrium" and proposes an approach based on Pareto optimization to identify the optimal collaborators. Besides, edge devices sometimes upload unreliable data either intentionally (the data poisoning attack) or unintentionally (low-quality data) \cite{9669031}, 
which may cause difficulties in model aggregation. \cite{kang2020reliable} introduces ``reputation" as a metric and proposes a reliable device selection scheme based on reputation to achieve rational collaboration. Another way to achieve rational collaboration is to motivate the clients to use high-quality data in the training process. \cite{zhang2022game} implements a game theoretic mechanism to reward the federated participants only when they utilize high-quality data, and adopt an equilibrium strategy to ensure that the reward for each client achieves its optimal value. Due to the heterogeneity of old and new tasks in FCL, the problem of rational collaboration becomes more complex and has a greater impact on model performance.

\textbf{Convergence Time}: Due to the need for multiple rounds of knowledge communication and model iteration, FL often implies longer convergence time \cite{bonawitz2019towards}. Some methods used for overcoming catastrophic forgetting in CL, such as rehearsal and generative replay, can result in slower convergence \cite{verwimp2021rehearsal}. As the combination of FL and CL, FCL still has room for improvement in this area. Theoretically, reducing convergence time is crucial for efficient FCL training, since it can enable quicker deployment of updated models and more efficient utilization of computational resources. And numerous rounds of communication can inevitably increase the risk of privacy breaches or security vulnerabilities because it extends the duration of the exposure of information and thus weakens the privacy and security guarantees of FCL. Besides, the users of edge nodes also expect a shorter convergence time. However, most of the existing works on FCL only focus on accuracy. It is necessary for researchers to take convergence time into account and develop new techniques to improve the efficiency of FCL algorithms. 

\textbf{Recommendation System}: \cite{huang2021feddsr} incorporates FL into the reinforcement learning model to provide a more secure daily schedule recommendation. \cite{zhou2019privacy} designs a distributed recommendation system via FL to enable a group of agents to collaboratively learn the tastes of various users. However, since user preferences and item catalog vary over time\cite{cai2022reloop}, a federated continual recommendation system needs to be deployed to provide up-to-date recommendations that align with the current preferences of users.

\textbf{Large Language Model}: The development of large language models (LLMs) has extensively promoted the application of artificial intelligence in various fields \cite{min2023recent}. However, the risk of privacy leakage deters some companies from adopting LLMs. To pre-train LLMs in specialized domains as well as protect data privacy, researchers integrate FL into LLMs. \cite{xinfedbfpt} proposes a communicational efficient FL framework for pre-training BERT by training
the shallower layers progressively. FedBERT \cite{tian2022fedbert} allows the clients to fine-tune the private NLP tasks individually after pre-training the global model. There are also some recent works exploring the use of LLMs to address the challenges in CL. \cite{wang2022learning} alleviates temporalCF by learning to prompt the pre-trained LLM dynamically. CPT \cite{ke2022continual} continually post-trains the LLMs with incremental unlabeled domain corpora to expand the knowledge of LLMs without forgetting. Post-training \cite{xu2019bert} refers to the technique that reduces bias introduced by non-review data and fuses domain knowledge into LLMs. \cite{ke2021achieving} inserts capsule layers into LLMs to encourage knowledge transfer among tasks and isolate task-specific knowledge. Generally, despite its practical significance, the research on combining FCL and LLM is still limited \cite{ke2022continualsurvey}.

\textbf{Other Industries} can also apply FCL techniques to their needs. \cite{sun2023federated} proposes a federated continual blockchain framework for physiological signal classification. \cite{schreyer2022federated} recently utilizes FCL to assist human auditors, since it can better satisfy the ``real-time" assessment of digital accounting journal entries. \cite{lanza2023urban} facilitates urban traffic forecasting with peer-to-peer FCL framework.

As stated above, FCL is still a newly emerging research area, with many promising avenues yet to be explored.

\section*{Acknowledgments}
This work was supported by the Natural Science Foundation of Sichuan Province (No. 2022NSFSC0528), the Sichuan Science and Technology Program (2022ZYD0113), and the Beijing Natural Science Foundation (4212021).

\bibliographystyle{IEEEtran}
\bibliography{references}

\vspace{-33pt}
\begin{IEEEbiography}[{
\includegraphics[width=1in,height=1.25in,clip,keepaspectratio]{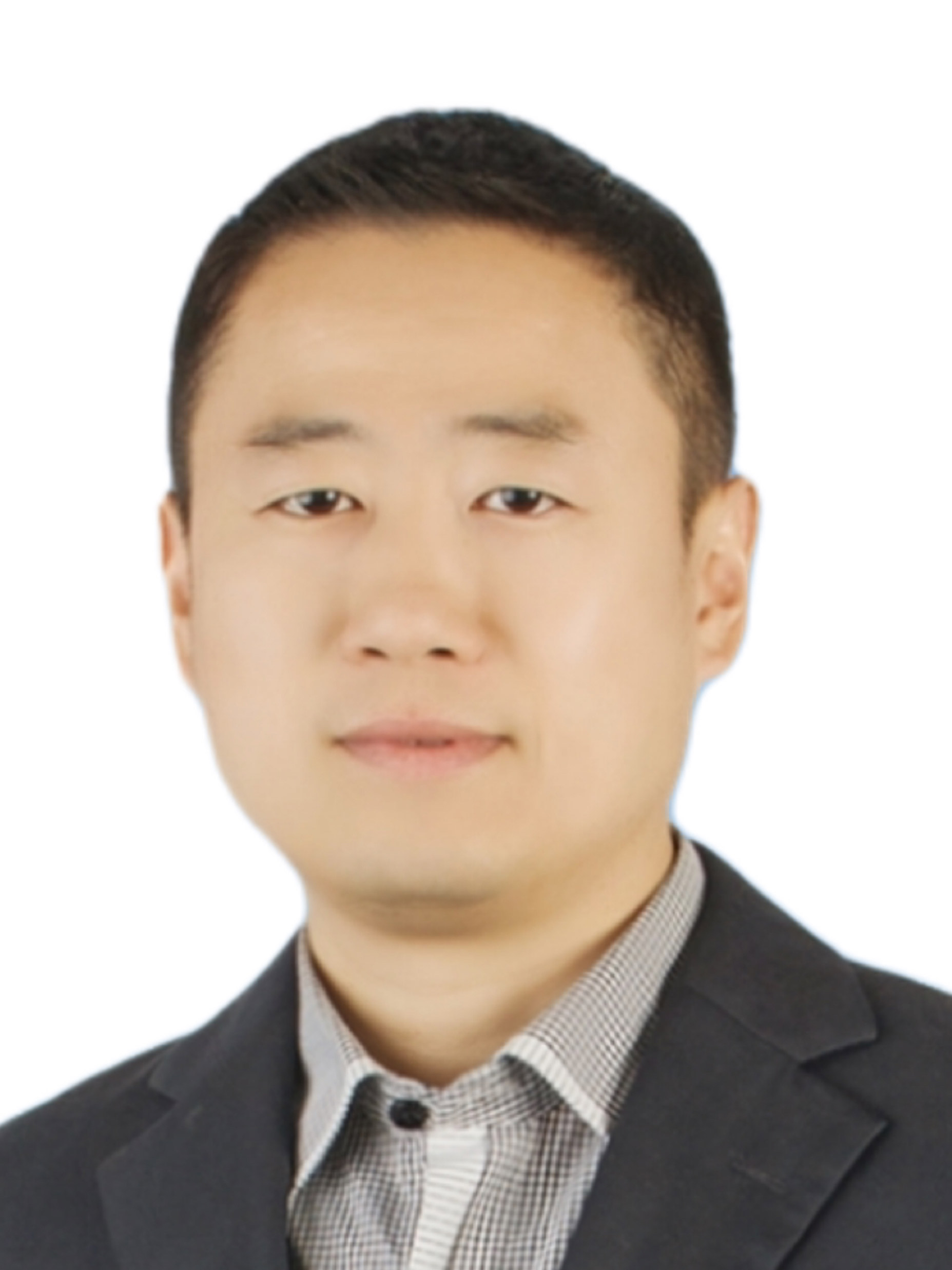}}]{Xin Yang} (Member, IEEE) received the Ph.D. degree in computer science from Southwest Jiaotong University, Chengdu. He is currently a Professor at the School of Computing and Artificial Intelligence, Southwestern University of Finance and Economics. He has authored more than 70 research papers in refereed journals and conferences. His research interests include machine learning, continual learning and multi-granularity learning.
 \end{IEEEbiography}

\vspace{-33pt}
\begin{IEEEbiography}[{
\includegraphics[width=1in,height=1.25in,clip,keepaspectratio]{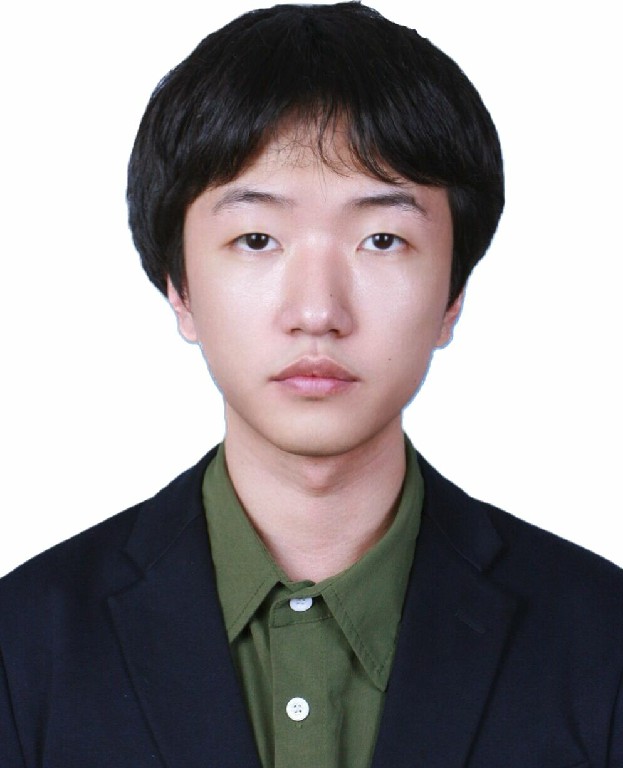}}]
{Hao Yu} received the B.S. degree from Southwest Petroleum University, Chengdu, China, in 2022. Now he is purchasing an M.S. degree from the School of Computing and Artificial Intelligence, Southwestern University of Finance and Economics. His main research interests include federated continual learning and continual learning.
\end{IEEEbiography}

\vspace{-33pt}
\begin{IEEEbiography}[{
\includegraphics[width=1in,height=1.25in,clip,keepaspectratio]{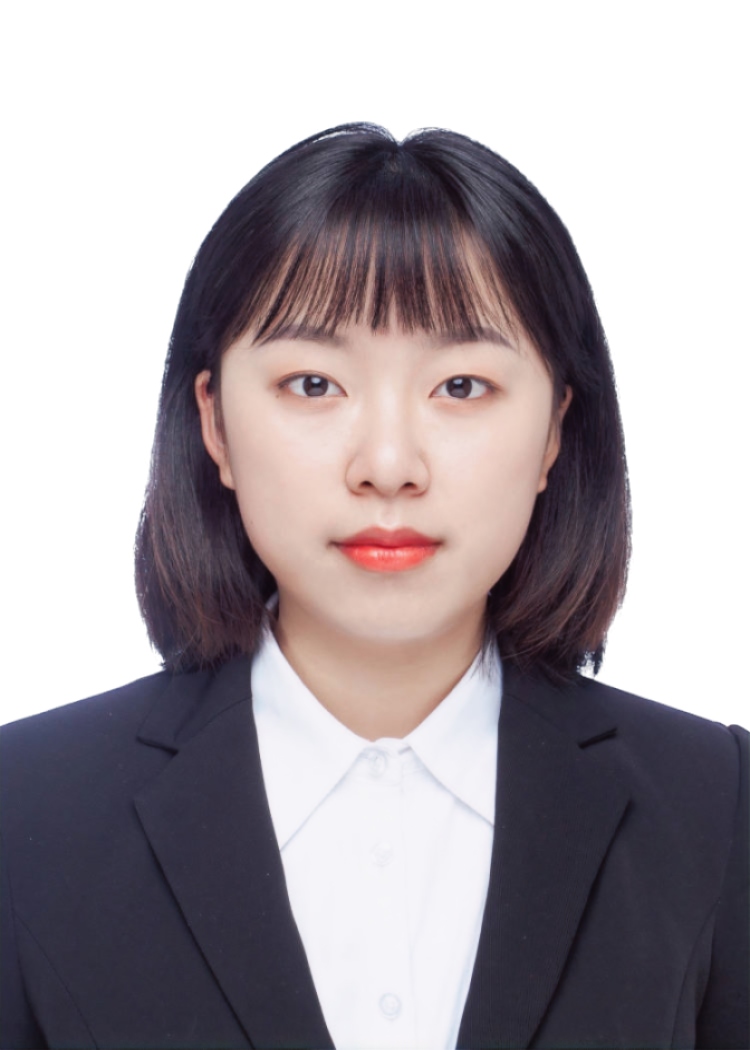}}]
{Xin Gao} received the B.S. degree from Sichuan University, Chengdu, China, in 2021. Now she is purchasing an M.S. degree from the School of Computing and Artificial Intelligence, Southwestern University of Finance and Economics. Her main research interests include federated continual learning and granular computing.
\end{IEEEbiography}

\vspace{-33pt}
\begin{IEEEbiography}[{
\includegraphics[width=1in,height=1.25in,clip,keepaspectratio]{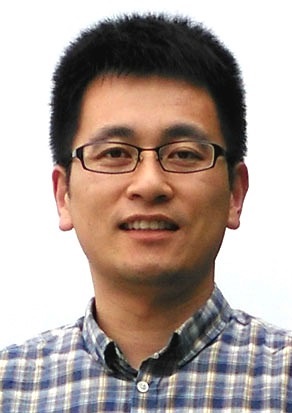}}]
{Hao Wang} is a Research Fellow at Nanyang Technological University (NTU), Singapore. He received his Ph.D. in Computer Science from Southwest Jiaotong University. Before joining NTU, he was with Zhejiang Lab. He was also a Visiting Student with the University of Illinois at Chicago (UIC) for two years. His research interests include lifelong and continual learning, sentiment analysis, multi-view learning, NLP, and spatio-temporal data mining.
\end{IEEEbiography}

\vspace{-33pt}
\begin{IEEEbiography}[{
\includegraphics[width=1in,height=1.25in,clip,keepaspectratio]{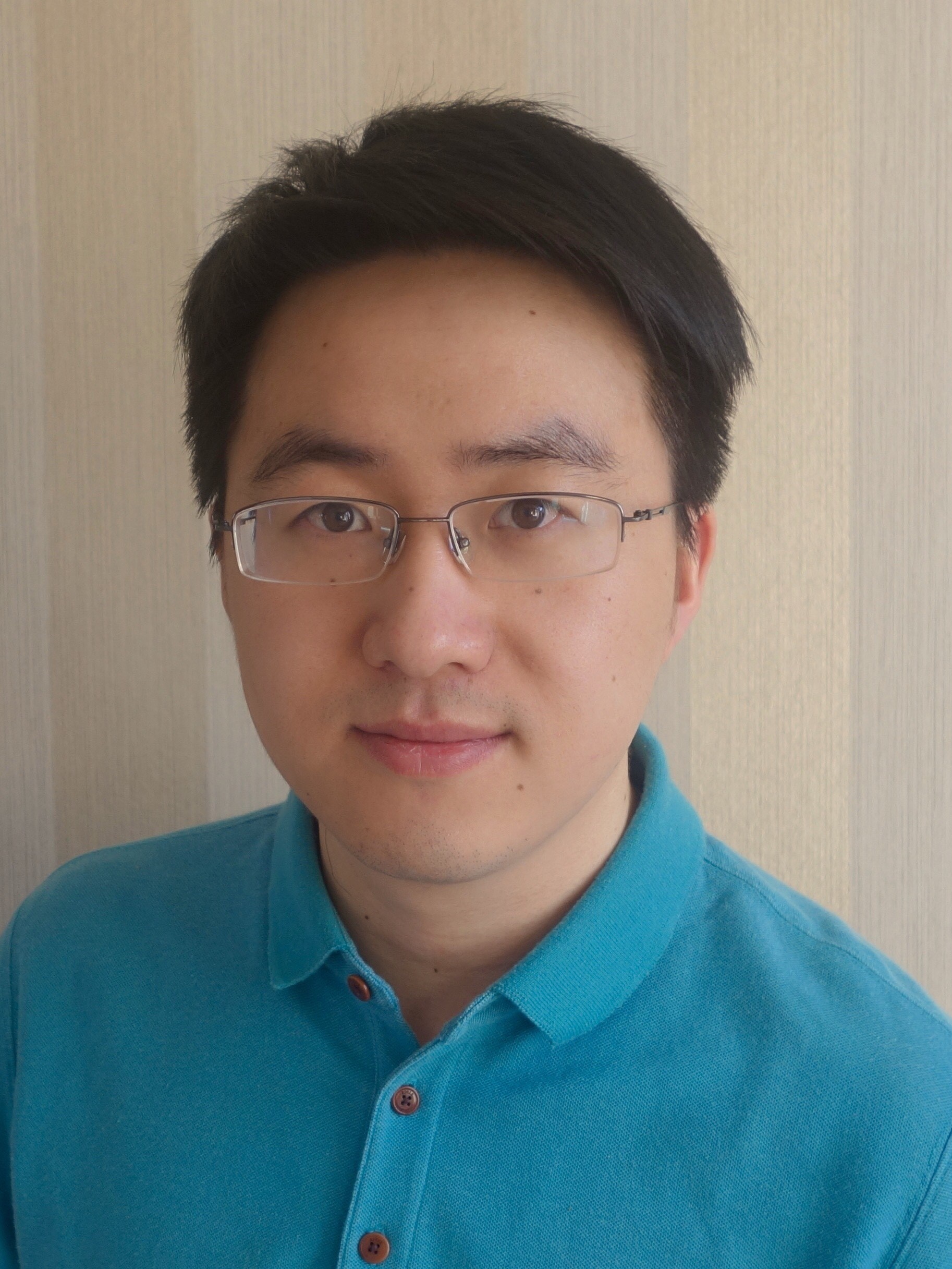}}]
{Junbo Zhang} (Senior Member, IEEE) is a Senior Researcher of JD Intelligent Cities Research. He is leading the Urban AI Product Department of JD iCity at JD Technology, as well as the AI Lab of JD Intelligent Cities Research. Prior to that, he was a researcher at Microsoft Research Asia (MSRA). He has published over 50 papers in spatio-temporal data mining and AI, urban computing, deep learning, and federated learning. He serves as an Associate Editor of ACM Transactions on Intelligent Systems and Technology. He received a number of honors, including the Second Prize of the Natural Science Award of the Ministry of Education in 2021, the 22nd China Patent Excellence Award in 2021, the ACM Chengdu Doctoral Dissertation Award in 2016, the Chinese Association for Artificial Intelligence (CAAI) Excellent Doctoral Dissertation Nomination Award in 2016. He is a senior member of CCF and IEEE, a member of ACM.
\end{IEEEbiography}

\vspace{-33pt}
\begin{IEEEbiography}[{
\includegraphics[width=1in,height=1.25in,clip,keepaspectratio]{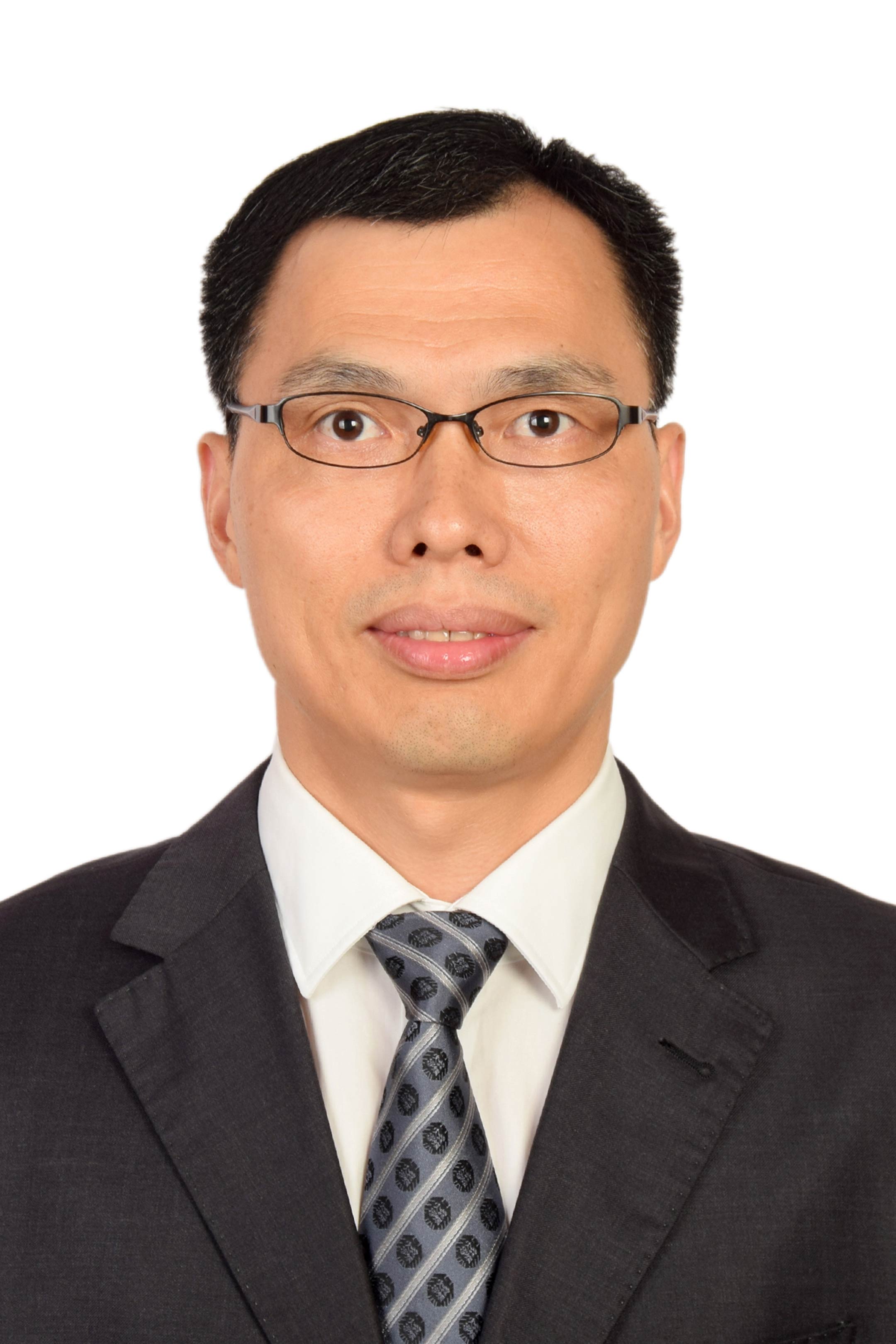}}]{Tianrui Li} (Senior Member, IEEE) received the B.S., M.S. and Ph.D. degrees from Southwest Jiaotong University, Chengdu, China, in 1992, 1995, and 2002, respectively. He was a Post-Doctoral Researcher with Belgian Nuclear Research Centre, Mol, Belgium, from 2005 to 2006, and a Visiting Professor with Hasselt University, Hasselt, Belgium, in 2008; the University of Technology, Sydney, Australia, in 2009; and the University of Regina, Regina, Canada, in 2014. He is currently a Professor and the Director of the Key Laboratory of Cloud Computing and Intelligent Techniques, Southwest Jiaotong University. He has authored or co-authored over 300 research papers in refereed journals and conferences. His research interests include big data, machine learning, data mining, granular computing, and rough sets.
\end{IEEEbiography}

\vfill
\end{document}